\documentclass[]{beingbeyond}
\usepackage{enumitem}
\usepackage[toc,page,header]{appendix}

\usepackage[utf8]{inputenc} 
\usepackage[T1]{fontenc}    
\usepackage{hyperref}       
\usepackage{url}            
\usepackage{array}          
\usepackage{booktabs}       
\usepackage{amsfonts}       
\usepackage{nicefrac}       
\usepackage{microtype}      
\usepackage{xcolor}         
\usepackage{xspace}
\usepackage{bm}
\usepackage{bbding}
\usepackage{bbm}
\usepackage{tabularx}
\usepackage{amssymb}
\usepackage{enumitem}
\usepackage{amsmath}
\usepackage{mathtools}
\usepackage{amsthm}
\usepackage{multirow}
\usepackage{makecell}
\usepackage{color}
\usepackage{colortbl}
\usepackage{adjustbox}
\usepackage{caption}
\usepackage{graphicx}
\usepackage{soul} 
\usepackage{pifont}
\usepackage{wrapfig}
\usepackage{multicol}

\usepackage{xspace}
\makeatletter
\DeclareRobustCommand\onedot{\futurelet\@let@token\@onedot}
\def\@onedot{\ifx\@let@token.\else.\null\fi\xspace}

\def\eg{\emph{e.g}\onedot}

\def\ie{\emph{i.e}\onedot}

\definecolor{BlockC}{gray}{0.98}  
\definecolor{BlockA}{RGB}{191,211,230}
\definecolor{BlockB}{RGB}{199,233,192}

\title{Being-H0.5: Scaling Human-Centric Robot Learning \\for Cross-Embodiment Generalization}

\author{\textbf{BeingBeyond Team}
\vspace{5mm}}

\webpage{\url{https://research.beingbeyond.com/being-h05}}

\firstfig[width=\linewidth][\textwidth]
  {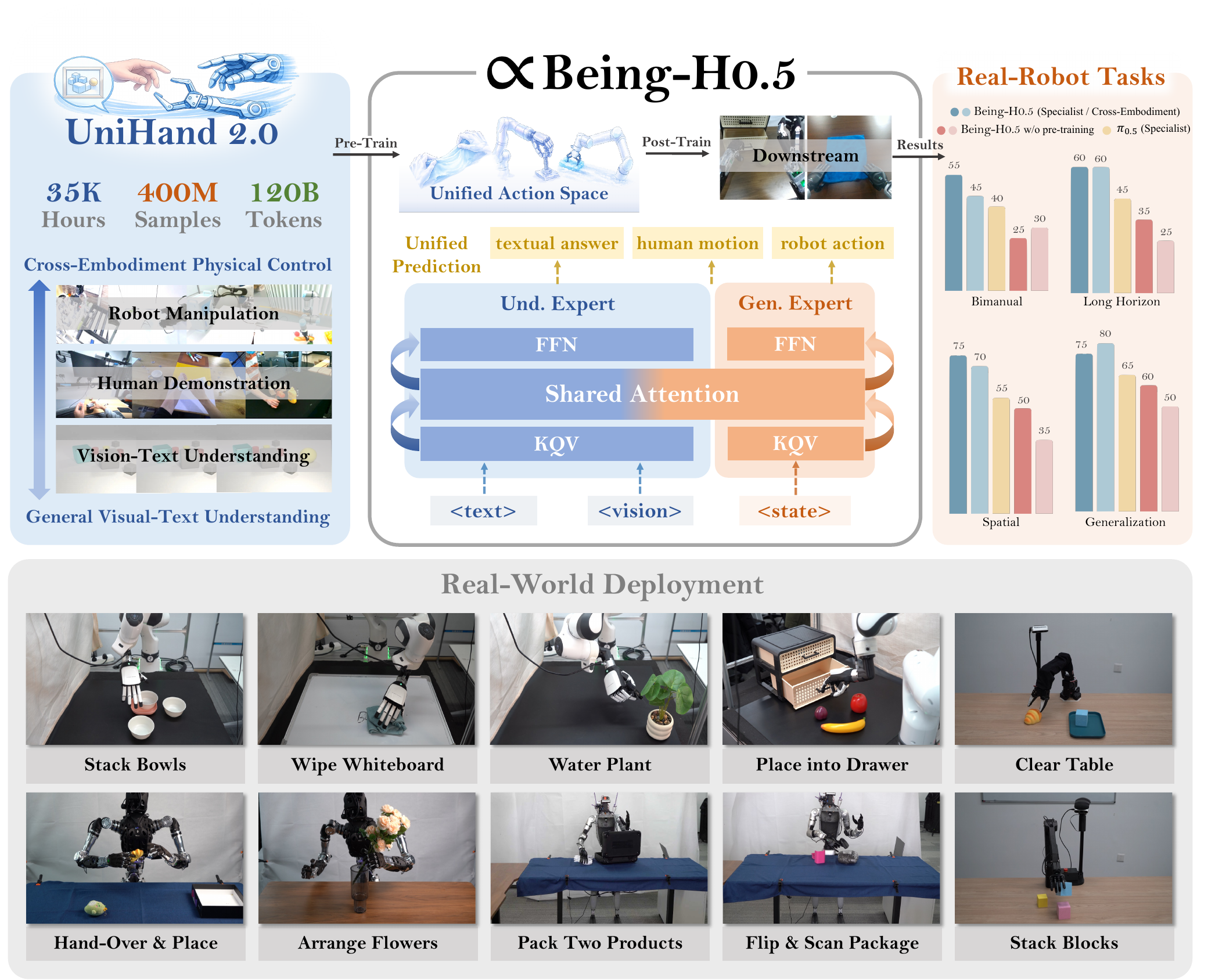}
  {\textbf{Being-H0.5 at a Glance.} We scale \textbf{human-centric robot learning} with Being-H0.5 toward cross-embodiment generalization. We introduce UniHand-2.0, a large-scale corpus exceeding \textbf{35,000 hours} that spans both cross-Embodiment physical control and general visual-text understanding. Building on this data, we unify human hand motion and diverse robot embodiments with a \textbf{Unified Action Space}, and train all heterogeneous supervision through \textbf{unified sequence modeling} under a single framework.  This yields a single foundation model that can \emph{perceive, describe, and act} within one framework, enabling robust cross-embodiment generalization and real-world deployment across diverse robots and tasks. We empirically deploy a \textit{single} checkpoint of Being-H0.5 to control PND Adam-U, Franka+Inspire, Unitree G1, BeingBeyond D1, and LeRobot SO-101 to accomplish diverse tasks.}
  {fig:first_fig_label}

\abstract{
We introduce Being-H0.5, a foundational Vision-Language-Action (VLA) model designed for robust cross-embodiment generalization across diverse robotic platforms. 
While existing VLAs often struggle with morphological heterogeneity and data scarcity, we propose a human-centric learning paradigm that treats human interaction traces as a universal “mother tongue” for physical interaction. 
To support this, we present UniHand-2.0, the largest embodied pre-training recipe to date, comprising over 35,000 hours of multimodal data across 30 distinct robotic embodiments. 
Our approach introduces a Unified Action Space that maps heterogeneous robot controls into semantically aligned slots, enabling low-resource robots to bootstrap skills from human data and high-resource platforms. 
Built upon this human-centric foundation, we design a unified sequential modeling and multi-task pre-training paradigm to bridge human demonstrations and robotic execution. 
Architecturally, Being-H0.5 utilizes a Mixture-of-Transformers design featuring a novel Mixture-of-Flow (MoF) framework to decouple shared motor primitives from specialized embodiment-specific experts. 
Finally, to make cross-embodiment policies stable in the real world, we introduce Manifold-Preserving Gating for robustness under sensory shift and Universal Async Chunking to universalize chunked control across embodiments with different latency and control profiles. 
We empirically demonstrate that Being-H0.5 achieves state-of-the-art results on simulated benchmarks such as LIBERO (98.9\%) and RoboCasa (53.9\%), meanwhile performing strong cross-embodiment capabilities on five robotic platforms.

}

\checkdata[Date]{Jan 20, 2026}

\begingroup

\endgroup

\begin{document}

\maketitle

\tableofcontents
\newpage

\section{Introduction}

True intelligence is characterized by the fluid expression of knowledge across diverse media.
In Natural Language Processing (NLP)~\cite{devlin2019bert, brown2020language}, multilingual pre-training~\cite{eisenschlos2019multifit, pires2019multilingual, conneau2019cross} has transcended linguistic boundaries by distilling a ``universal semantics'', facilitating seamless knowledge transfer from high-resource to low-resource languages.
Today, general-purpose robotics stands at a similar crossroads: the motor spaces of heterogeneous robots are, in essence, distinct physical languages.
Consequently, a primary challenge in scaling robotic intelligence is cross-embodiment translation:

\begin{center}
\begin{tcolorbox}[width=.9\textwidth, colframe=black!80, colback=gray!0, left=0.2cm, right=0.2cm, bottom=0.1cm, top=0.1cm]
How can robotic intelligence master a new platform with limited data, much like a human learning a minor language?
\end{tcolorbox} 
\end{center}

While Vision-Language-Action (VLA) models show promise toward a general robotic policy, they are hindered by a profound ``physical gap''.
Current VLAs function as monolingual speakers that are highly proficient on specific hardware but effectively incompatible when deployed on different morphologies.
This fragmentation is well-known exacerbated by the scarcity of robot-specific data~\cite{kim2024openvla, o2024openx}.
Unlike the trillions of tokens available in NLP, robotics lacks large-scale demonstration corpora for most individual platforms.
For diffusion-based VLAs~\cite{black2024pi_0, nvidia2025gr00t}, this is particularly acute: a model pre-trained on simpler hardware (e.g., parallel grippers) suffers from severe distribution shift especially when encountering the high-dimensional vector fields of complex entities (e.g., dexterous hands).
This mismatch causes inference trajectories to ``drift'' from the valid motion manifold, resulting in unstable behaviors. 

To overcome these hurdles, we propose a paradigm shift toward embodiment-agnostic VLAs via two core concepts:
\begin{itemize}[leftmargin=12pt]
    \item \textbf{Human-Centric Learning:}
    Just as different human languages share an underlying grammar (e.g., syntax and logic), different robots follow shared physical laws. 
    We treat the vast library of human interaction acts as the ``mother tongue''---the universal lingua franca of the physical world. 
    Human movements provide dense semantic priors, capturing the logic of causal interaction and the physics of contact that remain invariant across all kinematic ``dialects''. 
    By grounding learning in human-centric data, we provide a source of physical common sense that enables VLAs to adapt rapidly to new hardware.
    \item \textbf{Unified Action Space:}
    By representing diverse robot actions as tokens in a shared physical vocabulary, we align disparate robot morphologies into a unified latent space, allowing low-resource, complex robots to bootstrap motor skills from data-rich platforms and human demonstrations.
\end{itemize}

Building upon these motivations, we introduce \textbf{Being-H0.5}, a foundational VLA towards cross-embodiment generalization.
Our model is anchored by UniHand-2.0, an expansive corpus comprising \textbf{35,000+ hours of data and 120 billion tokens, totaling more than 400 million samples}.
As an extension of ~\cite{luo2025beingh0}, UniHand-2.0 integrates 16,000 hours of egocentric human video, 14,000 hours of robot manipulation, and 5,000 hours of general visual-language understanding data.
Notably, UniHand-2.0 spans \textbf{30 distinct robotic platforms}, from tabletop arms to legged humanoids.
Much like a multilingual LLM, Being-H0.5 is pre-trained on this ``polyglot'' dataset to internalize a robust, shared representation of physical manipulation.
Furthermore, UniHand-2.0 includes large-scale vision-text corpora to ensure the model retains the high-level reasoning, strategic planning, and instruction-following capabilities of its VLM backbone.
Compared to its predecessor, UniHand-2.0 represents a $200\times$ increase in scale.
To our knowledge, this represents the largest embodied pre-training recipe to date, featuring the most extensive use of human video and the most diverse collection of robotic embodiments.
By centralizing pre-training on massive human data alongside dozens of robotic datasets, we endow Being-H0.5 with strong generalization across heterogeneous platforms, even when target-specific data is highly limited.

Furthermore, we developed a human-centric data collection system named \textbf{UniCraftor} to address the limitations of existing datasets and scale our data acquisition. Through meticulous design, our system integrates depth information, keyframe events, and precise camera extrinsics—modalities that are often absent but important in current benchmarks. The system is intentionally modular and extensible, specifically designed to accommodate diverse and flexible viewpoints, as well as future tactile information. Leveraging this system, we have curated a comprehensive dataset exceeding 200 hours, covering 43 distinct tasks.

In addition to data scarcity, scaling generalist robotics is constrained by embodiment heterogeneity---the physical variance between human morphology and diverse robotic platforms with disparate kinematics, actuation limits, and control frequencies.
Naively aggregating these datasets introduces action-space interference, where conflicting control signals generate significant noise.
This often results in negative transfer, where cross-platform training degrades performance rather than fostering synergy.
Some works~\cite{nvidia2025gr00t} typically mitigate this by employing platform-specific action heads. 
However, this strategy is inherently limited as it bypasses the underlying structural misalignments.
Without reconciling these configurations, models fail to develop proprioceptive-aware reasoning or distill transferable physical priors across domains~\cite{zheng2025xvla}.
Consequently, existing policies remain either embodiment-specific specialists or ``shallow generalists'' that struggle with high-DoF tasks, such as dexterous or bimanual manipulation.
To bridge this gap, we propose a \textbf{Unified Action Space} that maps human trajectories and heterogeneous robot controls into semantically aligned slots.
This acts as a ``universal grammar'' for disparate hardware, effectively decoupling functional intent (e.g., the semantic objective of a grasp) from the mechanical articulation.
By doing so, our model internalizes the underlying physics of interaction rather than just embodiment-specific commands. 

Building on this unified action space, we take a further step towards scalable pre-training by \textbf{casting all heterogeneous supervision into a unified sequence modeling problem}.
Rather than maintaining separate pipelines for human demonstrations, robot trajectories, and vision-language corpora, we serialize them into a single multimodal token stream, where \emph{vision} and \emph{text} provide contextual grounding and unified \emph{state/action} tokens carry physically meaningful interaction signals.
This lets us optimize all data with various losses applied the relevant supervised segments. Concretely, text-centric corpora (e.g., VQA and motion description) contribute a standard next-token prediction loss over \texttt{text}, while human and robot corpora supervise \texttt{action} prediction in the unified space, with human behavior supplying dense, transferable behavioral priors and robot trajectories anchoring high-fidelity kinematic control.
By unifying both the representation and the optimization, pre-training becomes a coherent curriculum over one token stream, enabling the model to perceive, describe, and act within the same framework.

Architecturally, Being-H0.5 adopts a Mixture-of-Transformers (MoT) design that disentangles high-level multimodal reasoning from low-level execution experts.
While a unified action space effectively represents diverse embodiments and prevents representational conflicts during pre-training, the limited capacity of conventional action experts remains a critical bottleneck.
Inspired by the principle of Mixture-of-Experts (MoE)~\cite{shazeer2017moe}, we introduce a scalable architectural framework called \textbf{Mixture of Flow}, which decouples the action module into foundational experts with shared dynamic knowledge and specialized experts that utilize embodiment-aware task routing.
To ensure practical scalability, we introduce two complementary ingredients that maintain flow-based action generation both \emph{stable} and \emph{transferable}.
\textbf{1) Manifold-Preserving Gating} encourages the model to rely on reliable context and fall back to robust priors when perception is ambiguous, preventing unstable corrections from being amplified through iterative refinement.
\textbf{2) Universal Async Chunking} extends real-time chunked control to the \emph{cross-embodiment} setting.
Rather than baking in a single robot’s timing and control assumptions, we train a unified policy to remain consistent across heterogeneous platforms with varying actuation frequencies and delay profiles, enabling a single checkpoint to operate fluently on all embodiments.
Together, these innovations allow Being-H0.5 to remain fluent across varying sensory conditions and heterogeneous robot ``dialects''.

We evaluate Being-H0.5 and the efficacy of human-centric learning through extensive benchmarking across real-world and simulated environments.
Being-H0.5 achieves state-of-the-art (SoTA) results on LIBERO (98.9\%) and RoboCasa (53.9\%) using only low-resolution RGB input without auxiliary modalities.
Furthermore, our model significantly outperforms existing VLAs, such as $\pi$0.5, across five physically distinct embodiments, which demonstrates superior cross-embodiment generalization regardless of structural complexity.
\ul{Notably, our real-robot experiments reveal an unexpected emergent embodiment-level zero-shot transfer signal:
A single Being-H0.5 generalist checkpoint, trained jointly across embodiments under a unified action interface, achieves a non-zero success rate on unseen tasks ---embodiment pairs without any data on the target robot.}
We believe this observation sheds light on a practical scaling direction for cross-embodiment VLAs and paves the way for improving emergent transfer through increasing diverse post-training datasets.

To support the community and ensure full reproducibility, we provide a comprehensive open-source release encompassing our model weights, training pipeline, and simulation scripts.
Furthermore, we will include our real-world deployment infrastructure and a 1,000 GPU-hour pre-training recipe.
Our primary contributions are summarized as follows:
\begin{itemize}[leftmargin=12pt]
    \item \textbf{Largest Training Recipe.} 
    We introduce UniHand 2.0, the most extensive embodied VLA dataset to date, comprising 400M samples across 35,000 hours (16,000 human, 14,000 robot) and 30 embodiments. 
    To facilitate future scaling, we also provide a portable, plug-and-play human data collection system.
    \item \textbf{Unified Training Paradigm.}
    For the first time, we unify human hand motion and diverse robotic controls into a single action space for cross-embodiment generalization. 
    This is coupled with a unified sequence modeling paradigm that allows Being-H0.5 to \emph{perceive, describe, and act} within a single sequence, enabling scalable human-centric pre-training on heterogeneous corpora.
    \item \textbf{Architectural Innovations.} 
    To maximize model capacity and cross-domain transfer, we introduce several novel designs, including mixture-of-flow, manifold-preserving gating, and universal async chunking.
    These components address the inherent bottlenecks in scaling flow-based action generation.
    \item \textbf{Real-Time Infrastructure.} 
    We develop an efficient inference infrastructure that enables low-latency, real-time control, providing the computational throughput necessary for high-DoF, complex platforms.
    \item \textbf{State-of-the-Art Results.}
    Evaluated on five distinct real-world platforms and major simulated benchmarks, Being-H0.5 sets new SoTA records, including 98.9\% on LIBERO and 53.9\% on RoboCasa, while demonstrating emergent zero-shot transfer to unseen robot morphologies.
\end{itemize}

\section{Related Work}
\textbf{Vision-Language-Action Models.}
Recent advances in robotic manipulation~\cite{berscheid2019robot, dasari2019robonet, fang2023rh20t, shafiullah2023bringing} have shifted from narrow, single-task specialists toward generalist models trained on diverse, large-scale datasets spanning multiple scenes and embodiments.
A prominent paradigm in this shift is the Vision–Language–Action models (VLAs) ~\cite{brohan2022rt,zitkovich2023rt,kim2024openvla,janner2022planning,chi2025diffusion}, which bridges internet-scale perception and physical execution by fine-tuning pre-trained vision-language models (VLMs)~\cite{steiner2024paligemma, li2025eagle, wang2024qwen2vl, zhang2025bpe, zhang2025unified, hao2025openmmego, feng2025videoorion} on robotic control data.
While many VLAs share architectural backbones, they diverge significantly in action-head design.
Early initiatives typically adopt autoregressive approaches, discretizing continuous actions into tokens to align with VLM training objective~\cite{kim2024openvla, qu2025spatialvla}.
While this facilitates seamless knowledge transfer, it often introduces high inference latency and limits precision in high-degree-of-freedom tasks.
To mitigate these bottlenecks, recent frameworks~\cite{li2024cogact, janner2022planning, liu2024rdt, liang2025discrete} such as $\pi_0$~\cite{black2024pi_0} and GR00T-N1~\cite{nvidia2025gr00t} integrate diffusion-based architectures~\cite{ho2020denoising} (\eg, flow matching~\cite{lipman2022flow}) to generate high-frequency action chunks, which has been widely adopted by follow-up works~\cite{intelligence2025pi05, wen2025dexvla,zhong2025dexgraspvla}.
Despite these improvements, bridging the gap between high-level reasoning and granular execution remains a challenge.
Some approaches~\cite{zawalski2024ecot, lin2025onetwovla, clark2025action} employ Chain-of-Thought (CoT) reasoning~\cite{wei2022chain} to decompose long-horizon tasks such as OneTwoVLA’s adaptive ``thinking'' mode~\cite{lin2025onetwovla} or RAD’s language-based guides derived from human videos~\cite{clark2025action}, yet textual planning often lacks the spatial precision required for physical interaction.
Consequently, recent research has explored non-textual intermediate representations, such as bounding boxes~\cite{griffin2023mobile}, dense correspondence fields~\cite{laskin2020curl}, or 3D points~\cite{ten2017using}.
A notable example is MolmoAct~\cite{lee2025molmoact}, which transfers observations into depth-aware perception tokens to generate mid-level spatial plans as editable trajectory traces.
Despite these advances, most VLAs remain bottlenecked by their reliance on lab-collected teleoperation data, which lacks the environmental variety of the real world.
We address this by leveraging millions of egocentric videos to capture the behavioral richness necessary for robust generalization.

\textbf{Heterogeneous Pre-Training Datasets.}
The efficacy of VLAs is fundamentally tied to the scale and diversity of pre-training data.
Large-scale initiatives like Open X-Embodiment dataset~\cite{o2024openx}, along with datasets such as Droid~\cite{khazatsky2024droid} and BridgeData~\cite{walke2023bridgedata}, have aggregated thousands of hours of robot demonstrations across diverse platforms.
Community-driven benchmarks, such as~\cite{shukor2025smolvla}, have further expanded this landscape by leveraging accessible, portable hardware.
To address the complexities of high-dimensional bimanual coordination and dexterous manipulation, recent datasets like AgiBot World~\cite{bu2025agibot}, RoboMIND/2.0~\cite{wu2024robomind, hou2025robomind2}, and RoboCOIN~\cite{wu2025robocoin} have emerged to fill critical gaps in fine-grained motor control, while Open Galaxea~\cite{jiang2025galaxea} targets the unique requirements of mobile manipulation.
Despite these advances, most teleoperated data remains confined to specific lab settings, creating a bottleneck for generalizable policy learning.
While synthetic data~\cite{ye2025dex1b, chen2025internvla-m1} offers scalability, it suffers from the persistent sim-to-real gap.
Crucially, as these datasets span heterogeneous platforms with significant structural distinctions, existing VLAs lack a unified mechanism to integrate them effectively.
Discrepancies in data formats, annotation granularity, and embodiment structures pose significant challenges for large-scale unified modeling.
In this paper, we propose a unified state-action space to bridge these structural gaps.

\textbf{Human-Centric Learning.} 
The scarcity of real robot data, driven by the high cost of manual teleoperation, has prompted a shift toward human-centric learning.
To lower the barrier for data collection, portable physical interfaces like UMI~\cite{chi2024umi, xu2025dexumi, ha2024umi-leg} allow for the direct transfer of ``in-the-wild'' human skills to deployable robot policies.
Recent efforts have already scaled UMI-collected data to over 10,000 hours~\cite{genrobot2025realomin10kh}.
An even more scalable alternative is learning directly from abundant human videos.
Compared to robot-collected data, human video corpora span a vastly broader distribution of environments, object configurations, and long-horizon task structures.
This domain has evolved from traditional vision tasks such as recognition and grounding~\cite{feichtenhofer2019slowfast, lin2023univtg}, to large-scale egocentric benchmarks like Ego4D~\cite{grauman2022ego4d}, Ego-Exo4D~\cite{grauman2024egoexo4d}, EPIC-KITCHENS~\cite{damen2018epic}, and EgoDex~\cite{hoque2025egodex}.
While these provide massive interaction priors, the fundamental embodiment gap and the lack of task-aligned labels make their direct application to VLA modeling non-trivial.

Early efforts of human-centric learning attempted to exploit the visual diversity of human videos through representation learning~\cite{nair2022r3m,ma2022vip}.
However, the resulting features are largely implicit and lack action-level supervision, which limits their utility for precise downstream control.
Consequently, recent research has shifted toward more structured manners, which can be categorized into several trajectories.
First, some works~\cite{yang2025learning, chen2025villa, luo2025predictive} explore latent action as intent abstraction by encoding temporal changes into discrete codes or continuous embeddings to represent intent.
LAPA~\cite{ye2024latent} employs VQ-VAE tokens derived from frame pairs to pre-train a VLM, which is subsequently fine-tuned with an action head.
UniVLA~\cite{bu2025univla} replaces pixel reconstruction with DINOv2~\cite{oquab2023dinov2} feature prediction and introduces a two-stage codebook scheme to isolate task-centric latents, enabling autoregressive planning with lightweight latent-to-action decoding.
In addition to latent representation, approaches like GR-1/2~\cite{wu2023unleashing, cheang2024gr2} and Gen2Act~\cite{bharadhwaj2024gen2act} model physical interactions by explicitly predicting future environment states or video frames, treating video generation as a proxy for understanding dynamics.
A third line of work~\cite{chen2025vidbot,ma2025glover++,gavryushin2025maple,bahl2022human,bahl2023affordances} leverages explicit 2D geometric cues, such as point trajectories, keypoints, and bounding boxes~\cite{wen2023atm, yang2025magma, team2025gemini} to bridge the gap between video pixels and physical movement.

Moving beyond indirect proxies, some recent studies focus on recovering actual human actions from demonstrations to provide direct supervision for policy pre-training~\cite{kareer2025egomimic,feng2025vipa,punamiya2025egobridge,zhu2025emma, li2025scalable,bi2025h}. 
For instance, EgoVLA~\cite{yang2025egovla} constructs a shared action space using MANO hand-model parameters, retargeting 3D wrist and hand poses from egocentric videos to robot commands via inverse kinematics.
Being-H0~\cite{luo2025beingh0} scales this paradigm by introducing a motion tokenizer that discretizes continuous MANO trajectories into tokens for large-scale instruction tuning.
Despite these advancements, existing methods are often constrained by the difficulty of scaling high-quality action annotations and the lack of a framework that can seamlessly integrate heterogeneous human demonstrations with robot-collected data.
In this work, we extend the paradigm established by Being-H0 to curate the largest human-centric pre-training corpus to date, supported by a unified paradigm for multi-source integration.

\begin{figure}[t]
        \centering
        \includegraphics[width=\linewidth]{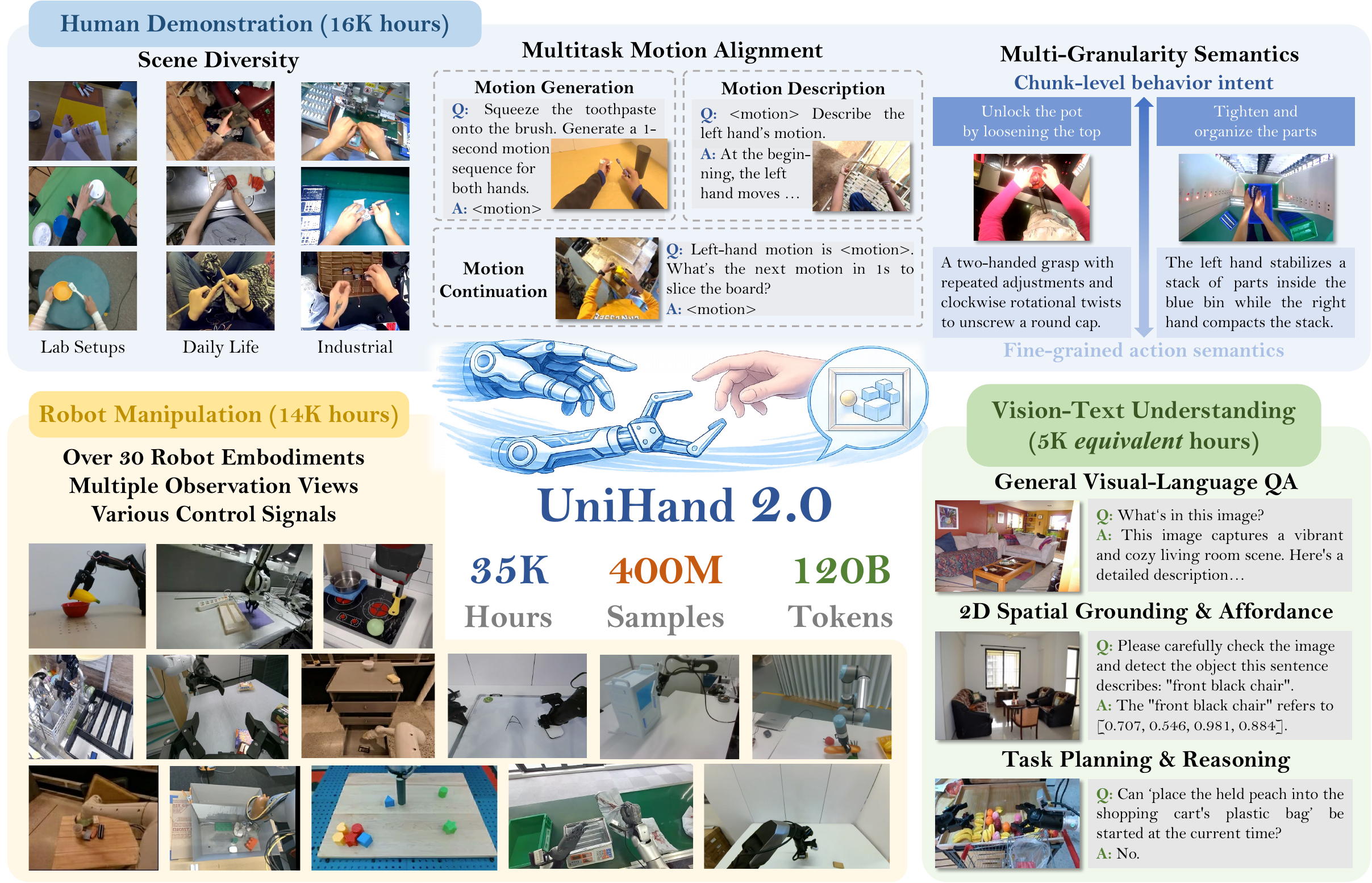}
        \caption{\textbf{Overview of UniHand 2.0.} UniHand 2.0 is our large-scale pre-training recipe for human-centric robot learning, comprising 35K hours of multimodal data from three complementary sources. 1) \textbf{Human Demonstration} with diverse scenes, multitask for motion alignment, and multi-granularity semantics. 2) \textbf{Robot Manipulation} spanning 30+ embodiments with multiple observation views and heterogeneous control signals. 3) \textbf{Vision–Text Understanding} covering general VQA, 2D spatial grounding \& affordance, and task planning \& reasoning.}
        \label{fig:dataset_overview}
\end{figure}

\section{UniHand-2.0: A Recipe for Large-Scale Human-Centric Learning} 
 
Vision-Language-Action models (VLAs) require vast quantities of robotic interaction data to acquire actionable knowledge and world commonsense.
However, the current landscape of robotic data is hindered by several fundamental limitations.
Our work systematically addresses the following critical challenges:

\begin{enumerate}[label=\textbf{\arabic*)}]
    \item \textbf{Limited Data Scale and Diversity.} 
    Most existing VLAs~\cite{kim2024openvla, kim2025openvla-oft} are constrained by the insufficient scale and diversity of their pre-training corpora. Many rely on a narrow scope of data, such as the Open X-Embodiment dataset~\cite{o2024openx}) which offers only six major subsets with restricted variety after rigorous filtering, while Agibot World~\cite{bu2025agibot} contains merely $\sim$200 hours of tabletop manipulation data and lacks essential third-person camera views.
    This scarcity impedes generalization to novel tasks and dynamic environments.
    To overcome this, we curate a comprehensive robotic manipulation dataset that aggregates the vast majority of available robot data.
    
    \item \textbf{Restricted Embodiment Variety.} 
    Beyond scale, existing datasets typically feature a limited suite of embodiments, often restricted to a single robot type (\eg, Agibot World~\cite{bu2025agibot} and LET~\cite{leju2025let}). 
    While some cross-embodiment datasets exist~\cite{wu2025robocoin}, unifying these sources for pre-training remains difficult due to significant structural discrepancies. 
    Consequently, few VLAs successfully incorporate broad robot morphologies. For example, the $\pi$-series~\cite{black2024pi_0, intelligence2025pi05} encompasses only 10 robot types, predominantly homogeneous bimanual platforms. 
    In contrast, UniHand-2.0 incorporates data from \textbf{30 distinct embodiments}, spanning single/dual-arm, portable, half-humanoid, and legged humanoid robots. 
    This represents the most extensive array of embodiments reported in VLA literature to date. 
    We achieve this by projecting these heterogeneous sources into a unified state-action space to ensure training stability.

    \item \textbf{Scarcity of Dexterous Hand Data.}
    Despite advances in general data collection, data involving dexterous hands remains exceptionally rare, comprising less than $5\%$ of existing corpora. 
    This inadequacy stems from the high cost of hardware and the low throughput of dexterous teleoperation. 
    To mitigate this, we leverage large-scale human motion data similar to ~\cite{luo2025beingh0} as a scalable proxy, capitalizing on the relative ease of capturing human hand interactions in natural settings.

    \item \textbf{Imbalance between Visual and Language Information.} 
    Prior VLAs~\cite{black2024pi_0, zhai2025wall-oss} generally rely on robot-only data for pre-training, leading to a severe modality imbalance where the ratio of text to visual tokens can reach 1:3,000. 
    This disparity causes models to lose substantial textual reasoning capabilities, which is critical for long-horizon task execution.
    While some works incorporate vision-language data~\cite{intelligence2025pi05, chen2025internvla-m1} or interleaved multimodal samples~\cite{qu2025eo1}, we systematically incorporate massive multimodal data during pre-training to ensure Being-H0.5 excels at both atomic action execution and long-horizon task planning \& spatial reasoning.

\end{enumerate}

\begin{figure}[h] 
\centering
\includegraphics[width=1\linewidth]{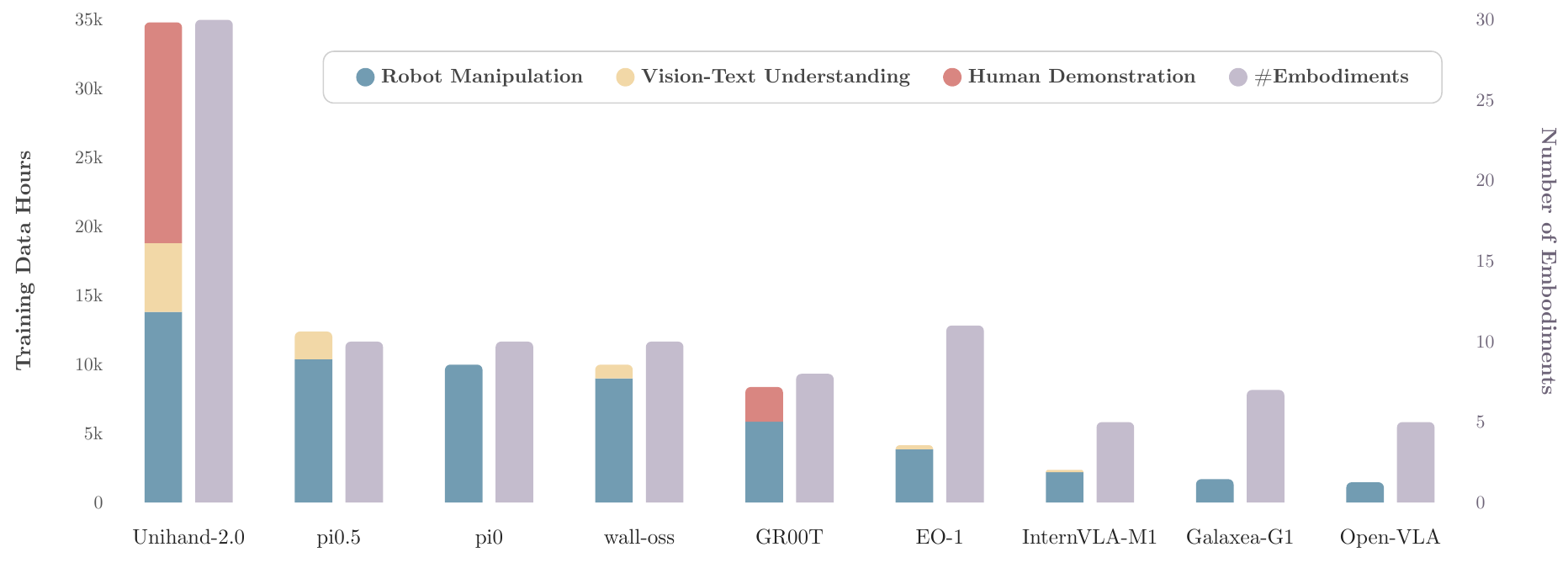}
\caption{\textbf{Comparison of training scale and embodiment diversity}. Stacked bars represent hours of training data (left axis); hatched bars represent embodiment counts (right axis).
UniHand-2.0 represents the largest and most diverse VLA pre-training recipe to date, totaling 35,000 hours of multimodal data.
This includes 16,000 hours of human data, 14,000 hours of robot data across 30 embodiments, and 5,000 equivalent hours of VLM data.} 
\label{fig:data_scale_compare}
\end{figure}

Here, we introduce \textbf{UniHand-2.0}, a significantly expanded dataset built upon its predecessor, UniHand-1.0~\cite{luo2025beingh0}. 
The dataset (Figure~\ref{fig:dataset_overview}) contains over \textbf{400 million} samples extracted from \textbf{35,000 hours} of multimodal data, totaling over 120B training tokens.
These samples span three key domains: egocentric human motion, robot manipulation, and visual-language understanding.
As illustrated in Figure~\ref{fig:data_scale_compare}, UniHand-2.0 is the largest human-centric VLA pre-training corpus to our knowledge.
Note that the usage of the Open X-Embodimen dataset is ambiguous for different works. We thereby restrict our embodiment counting with >10 hours of data.
UniHand-2.0 leverages low-cost human data as primary pre-training material, treating the human hand as a universal template for all end-effectors, imbuing models with foundational interaction knowledge and physics understanding. 
We advance this paradigm by extracting 134 million human data samples from \textbf{16,000 hours} of egocentric video, which is a $100\times$ increase over UniHand-1.0.
Furthermore, UniHand-2.0 incorporates over \textbf{14,000 hours} of diverse robotic data across \textbf{30 embodiment types} (\eg, Franka, AgiBot-GR1, Unitree-G1, SO101), endowing the model with robust cross-embodiment generalization.

We posit that such embodiment diversity is a prerequisite for existing diffusion-based VLAs.
From a manifold learning perspective, simple embodiments (\eg, parallel grippers) operate on a low-dimensional, smooth action manifold.
In contrast, dexterous robots inhabit complex, high-dimensional spaces where the manifold structure is often non-linear and fragmented.
The action distribution of these complex entities differs fundamentally from that of simple ones.
For instance, the binary ``open/close'' command of a gripper is a simple scalar, whereas the ``precision pinch'' of a dexterous hand requires a high-dimensional, coordinated vector.
This disparity leads to a severe target distribution shift during adaptation.
Furthermore, for a diffusion framework, it must infer a continuous vector field to define the probabilistic evolution toward the next action state. 
When a model pretrained only on simple robots encounters the unseen state space of a complex entity, its vector field predictions suffer from accumulated errors.
These errors cause generated trajectories to ``drift'' and rapidly deviate from the valid robot motion manifold, resulting in unstable or physically infeasible behaviors.

\begin{figure}[h] 
\centering
\includegraphics[width=1\linewidth]{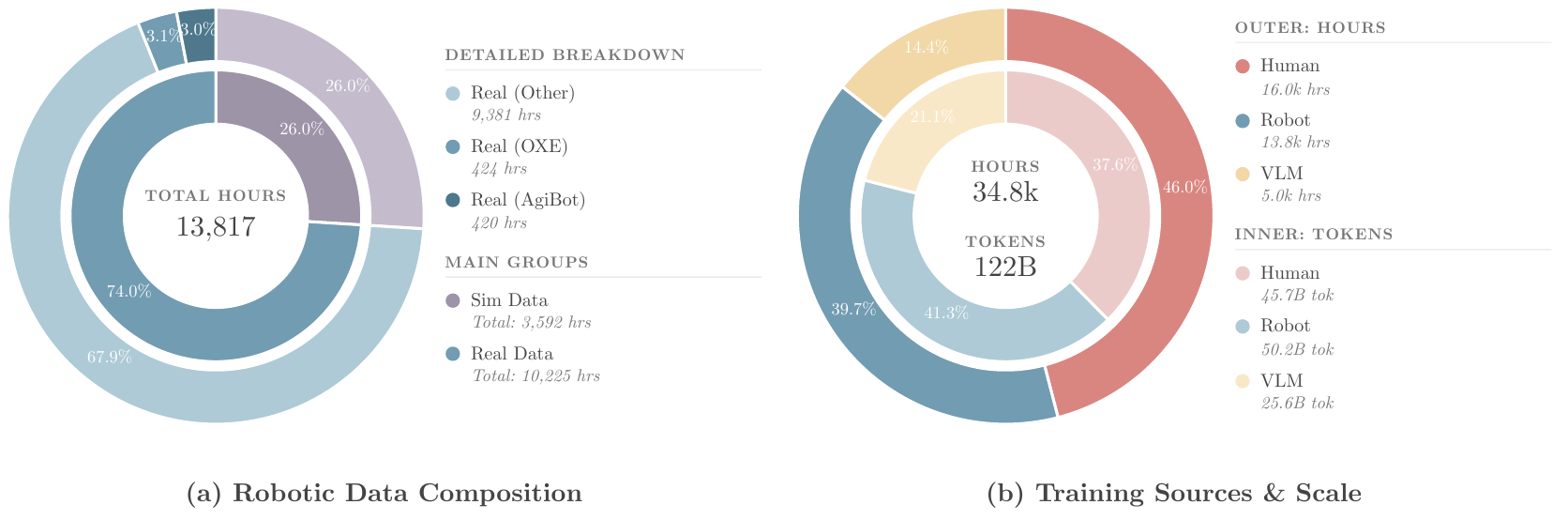}
\caption{\textbf{Statistics of UniHand-2.0. 
(Left) Ratio of simulation vs. real-world data}. 
We maintain a balanced ratio with simulation data at 26\%, while the widely used Open X-Embodiment (OXE) and AgiBot World datasets account for 3.1\% and 3.0\%, respectively.
\textbf{(Right) Training sources and scale}. Human, robot, and visual-language (``VLM'') data consist of 16K hours (25.6B tokens), 14K hours (45.7B tokens), and 5K equivalent hours (50.2B tokens), respectively.
These sources are curated to maintain a comparable scale for balanced pretraining.
}
\label{fig:dataset_statis}
\end{figure}

To maintain a critical balance between modalities, we integrate visual-language understanding data at a comparable scale, preserving the model’s reasoning and instruction-following abilities.
UniHand-2.0 thus constitutes the largest pre-training dataset organized under the human-centric learning concept. 
The following sections detail each component, and statistics are visualized in Figure~\ref{fig:dataset_statis}.

\subsection{Human Demonstration Data} 
Human demonstration data has emerged as a powerful and scalable foundation for pre-training VLAs, as evidenced by recent advancements~\cite{luo2025beingh0, yang2025egovla, nvidia2025gr00t, li2025scalable, fu2025metis}. 
Beyond serving as a substitute for scarce robotic interaction data, human-centric videos offer a direct window into the fundamental ways humans perceive, reason, and manipulate within the physical world.
These corpora capture a vast diversity of everyday activities---ranging from fine-grained hand-object interactions to complex tool use and rich contextual scenes---that would be prohibitively expensive to replicate at scale using physical robots. 
This diversity establishes robust, generalizable behavioral priors and provides a broad coverage of motion patterns and affordance that are often absent in scripted robotic collection or simulation environments. 

Following the paradigm established by Being-H0, we treat human hand motion as a generalized manipulator template. 
This paradigm physically grounds visual observations in real-world interactions, effectively transforming passive video frames into action-relevant supervision.
With hand motion as a grounding anchor, our model leverages egocentric human videos for robotic policy learning through three primary mechanisms:
\textbf{1) Action-Relevant Visual Context:} 
Hand motion functions as a mechanism for spatial attention, localizing where and what is relevant for a given action. This expands behavior-grounded visual context significantly beyond what is available in standard VLM corpora.
\textbf{2) Rich Manipulation Structure:}
Hand-centric interaction traces expose fine-grained contact patterns, dexterous control strategies, and diverse object affordance, capturing manipulation dynamics that are currently missing from most robot datasets.
\textbf{3) Diverse Textual Context for Intent Grounding:} 
By aligning linguistic context with hand-centric interaction traces, the model can ground human language in physical outcomes and motion, enabling intent-level learning rather than superficial visual pattern matching.

Building upon Being-H0, we significantly scale and diversify our human-video corpus, extending its scope beyond lab-recorded, manipulation-specific settings to include both open-source repositories and self-curated data.
To render these videos directly usable for robot learning, we extracted a unified hand-motion representation from diverse videos sources.
This is further supplemented by a high-fidelity in-house dataset to provide the signal precision often lacking in open-web videos.

\noindent\textbf{Curation of Heterogeneous Human Videos.}
In addition to the UniHand-1.0 dataset, we integrate an extensive collection of in-the-wild egocentric videos from large-scale public repositories, including Ego4D~\cite{grauman2022ego4d}, EPIC-KITCHENS~\cite{damen2020epic}, Egocentric-10K~\cite{buildai2025egocentric10k}, etc. 
This extension drastically increases environmental variety and embodied behavior, encompassing household chores, cooking, and industrial operations (see Figure~\ref{fig:dataset_overview}). 
To align these heterogeneous sources into a unified interface, we employ HaWoR~\cite{zhang2025hawor} to estimate hand poses via MANO parameters~\cite{romero2022mano} and camera extrinsics.
Furthermore, we enrich the dataset with fine-grained semantic supervision using Gemini-2.5~\cite{comanici2025gemini2.5}, generating dual-level annotations: detailed per-second instructions and holistic, 10-second chunk-level intent descriptions.

We structure the training data into three interconnected task families: \textbf{motion generation} (the primary objective mapping vision/language to kinematic action), \textbf{motion description}, and \textbf{motion continuation}. 
To ensure high data fidelity and minimize spurious correlations, we implement a rigorous four-stage post-processing pipeline:
\textbf{1) Language Augmentation:} 
We employ LLMs to paraphrase and diversify all template-formatted instructional texts, preventing the model from overfitting to rigid linguistic patterns.
\textbf{2) Motion-Quality Filtering:} 
We ensure data reliability by filtering noisy HaWoR estimates based on detection confidence and DBA error, while removing high-frequency jitter or discontinuities in wrist-space.
\textbf{3) Manipulation Relevance Filtering:} 
Using Gemini-assisted semantic screening, we exclude segments dominated by non-manipulative actions, such as pure locomotion.
\textbf{4) Handedness Debiasing:} 
We apply left-right spatial mirroring to all samples to mitigate common right-handed bias, thereby promoting ambidextrous generalization in the learned policy.

\begin{table*}[h!]
\centering
\small 
\setlength{\tabcolsep}{6pt} 
\renewcommand{\arraystretch}{1.05} 
 
\caption{\textbf{Statistical overview of robot manipulation data in UniHand-2.0}. 
The corpus aggregates demonstrations across \textbf{30 distinct embodiments}, totaling \textbf{14,000 hours} of interaction.
The dataset is characterized by high diversity in camera views, kinematic structures, and operational environments, categorized by end-effector type---``Grp'' (parallel gripper) vs. ``Dex'' (dexterous hand), and data source---``Real'' (physical world) vs. ``Sim'' (simulation).}
\label{tab:robot_data}
 
\begin{tabular}{@{}l l c c r@{}}
\toprule
\textbf{Robot Type} & \textbf{Camera Views} & \textbf{EEF Type} & \textbf{Source} & \textbf{Hours} \\ 
\midrule
 
\multicolumn{5}{l}{\textit{\textbf{Single-Arm Robots}}} \\ 
\addlinespace[0.2em] 
Franka        & ego, 3rd$\times$2, wrist & Grp/Dex & Real/Sim & 2196.4 \\ 
Google Robot  & ego, 3rd                 & Grp     & Real/Sim & 1195.2 \\ 
Kuka-iiwa     & ego, 3rd                 & Grp     & Real/Sim & 338.2 \\ 
WidowX        & 3rd$\times$2, wrist      & Grp     & Real/Sim & 305.2 \\ 
Xarm7 Robot   & top, 3rd                 & Grp     & Real/Sim & 594.4 \\ 
UR5E           & 3rd, wrist               & Grp     & Real/Sim & 208.3 \\ 
Hello Stretch & ego, 3rd                 & Grp     & Real     & 270.3 \\ 
Kinova Jaco          & 3rd$\times$3             & Grp     & Real/Sim & 194.6 \\ 
Kinova Gen3       & 3rd$\times$3             & Grp     & Sim      & 197.4 \\ 
Sawyer        & 3rd                      & Grp     & Sim      & 196.1 \\ 
Cobotta       & 3rd                      & Grp     & Real     & 10.6 \\ 
DLR Sara      & 3rd                      & Grp     & Real     & 12.1 \\ 
Willow Garage PR2           & ego                      & Grp     & Real     & 11.9 \\ 
RMC RM65           & 3rd                      & Dex     & Real     & 41.3 \\ 

\addlinespace[0.8em] 
\multicolumn{5}{l}{\textit{\textbf{Dual-Arm Robots}}} \\ 
\addlinespace[0.2em]
RMC Aida L         & high, wrist$\times$2 & Grp & Real     & 325.7 \\ 

Galaxea R1 Lite    & high, wrist$\times$2 & Grp & Real     & 630.5 \\ 
Agilex Split ALOHA & high, wrist$\times$2 & Grp & Real/Sim & 1099.1 \\ 
Agilex Cobot Magic & high, wrist$\times$2 & Grp & Real     & 399.2 \\ 
Piper              & ego, wrist$\times$2  & Grp & Real/Sim & 904.5 \\ 
 
\addlinespace[0.8em]
\multicolumn{5}{l}{\textit{\textbf{Portable Education Arm}}} \\ 
\addlinespace[0.2em]
BeingBeyond D1 & ego           & Dex & Real & 100.0 \\ 
SO101          & ego, wrist$\times$2 & Grp & Real & 194.4 \\ 
 
\addlinespace[0.8em]
\multicolumn{5}{l}{\textit{\textbf{Half-Humanoid}}} \\ 
\addlinespace[0.2em]
PND AdamU        & ego                 & Dex     & Real     & 200.0 \\ 
Agibot-G1       & \{ego, wrist\}$\times$2 & Grp/Dex & Real/Sim & 2391.7 \\ 
Leju kuavo 4 LB & ego, wrist$\times$2     & Grp/Dex & Real     & 1198.2 \\ 
AlphaBot 2      & ego, wrist$\times$2     & Grp     & Real     & 13.8 \\ 
Airbot MMK2     & 3rd, wrist$\times$2     & Grp     & Real     & 23.1 \\ 
Galbot G1       & wrist$\times$2          & Grp     & Sim      & 195.8 \\ 
Tianqing A2     & high, wrist$\times$2    & Grp     & Real     & 19.4 \\ 
 
\addlinespace[0.8em]
\multicolumn{5}{l}{\textit{\textbf{Humanoid}}} \\ 
\addlinespace[0.2em]
Unitree G1edu-u3 & ego, wrist$\times$2 & Grp/Dex & Real & 135.7 \\ 
Tiankung         & top                 & Grp     & Real & 214.3 \\ 
 
\midrule[\heavyrulewidth] 
 
\multicolumn{4}{l}{\textbf{Total: 30 embodiments}} & \textbf{13817.4} \\ 
\bottomrule
\end{tabular}
\end{table*}

\subsection{Robot Manipulation Data}
Robot control data serves as the cornerstone of policy learning, providing a critical pathway toward achieving both broad embodiment generalization and dexterous manipulation proficiency.
To establish a comprehensive resource for this purpose, we curate a large-scale, heterogeneous collection of robot demonstrations.
This dataset aggregates about 14,000 hours of interaction data (approximately 1.5 billion frames), making it one of the most extensive repositories of robotic behavior currently available.
The collection systematically integrates a wide array of datasets, including OpenX-Embodiment~\cite{o2024openx}, AgiBot-World~\cite{bu2025agibot}, SO100-Community~\cite{shukor2025smolvla}, InternData-M1~\cite{chen2025internvla-m1}, RoboMIND~\cite{wu2024robomind}, RoboCOIN~\cite{wu2025robocoin}, LET~\cite{leju2025let}, etc.
From this aggregate pool, we perform deduplication and downsample frames to 30\% to maximize data diversity while minimizing redundancy. 

As a result, our processed dataset encompasses \textbf{30 distinct robot embodiments}, such as Franka arms, Split ALOHA, and Agibot G1, representing a significant breadth of hardware morphologies and control interfaces. 
To further scale task coverage, we incorporate procedurally generated simulated samples~\cite{chen2025internvla-m1} and inpainting-augmented trajectories~\cite{ji2025oxe-auge}. 
While simulated data effectively boosts performance on standard benchmarks, excessive reliance can exacerbate the sim-to-real gap.
To mitigate this, we strictly cap the proportion of simulated data in the pretraining mixture (26\% of the total corpus, as shown in Figure~\ref{fig:dataset_statis}), ensuring that real-world signals remain the dominant learning signal.
Finally, we contribute novel demonstrations for hardware platforms such as the PND Adam-U and BeingBeyond D1 to improve data diversity and balance. 
Additional details regarding the robot data distribution are provided in Table~\ref{tab:robot_data}.

\subsection{Visual-Text Understanding Data}
A pronounced imbalance often exists between the visual and textual components of standard robotic training corpora.
While robotic interaction data provides dense visual signals, the accompanying linguistic supervision is frequently sparse and simplistic.
This disparity creates a fundamental asymmetry in multimodal supervision, risking a degradation in the model's capacity for complex textual reasoning, instruction following, and high-level planning.
For instance, within our robotic interaction corpus alone, visual tokens number 45.7 billion, whereas textual tokens amount to merely 30 million---a staggering ratio of approximately 1,000:1.
Left unaddressed, this discrepancy may cause the model to overfit to visual patterns at the expense of linguistic intelligence, potentially reducing a robotic agent to a visual-motor reflex system rather than a reasoning agent.
To restore modality balance, we curate a comprehensive visual-language understanding corpus specifically designed to preserve and enhance the model’s semantic reasoning capabilities.
We strategically categorize this collection into three pillars, ensuring coverage from general world understanding to precise, actionable robotic reasoning:

\begin{enumerate}[label=\textbf{\arabic*)}]

    \item \textbf{General Visual-Language QA. }
    To maintain robust general-purpose vision-language alignment, we incorporate established large-scale instruction-tuning datasets. 
    This includes image-based benchmarks such as LLaVA-v1.5~\cite{liu2024qwen-v1.5}, LLaVA-OneVision~\cite{li2024llava-ov} and FineVision~\cite{wiedmann2025finevision}, as well as video-centric datasets like LLaVA-Video~\cite{zhang2024llava-video}.
    These datasets serve to prevent the catastrophic forgetting of general world knowledge during robot-specific fine-tuning.
    
    \item \textbf{2D Spatial Grounding \& Affordance.}
    Robotic manipulation requires more than semantic description; it demands precise spatial localization. 
    To bridge the gap between semantic understanding and physical actuation, we integrate datasets focused on grounded reasoning, object localization, and affordance detection.
    Sources include RefCOCO~\cite{yu2016refcoco} and RefSpatial~\cite{zhou2025roborefer} for referring expression comprehension, alongside robotics-specific grounding datasets such as RoboPoint~\cite{yuan2024robopoint}, ShareRobot~\cite{ji2025robobrain}, RoboRefit~\cite{lu2023vl}, RoboVQA~\cite{sermanet2024robovqa}, MolmoAct~\cite{lee2025molmoact}, and A0-ManiSkill~\cite{xu2025a0}. 
    We also include point-level supervision benchmarks like PixMo-Points~\cite{deitke2025molmo-pixmo} and AsV2~\cite{wang2024asv2} to enhance the model’s fine-grained spatial awareness within the 2D image plane.

    \item \textbf{Task Planning \& Reasoning.} 
    In addition to immediate perception, a capable agent must reason over long horizons.
    We address this by incorporating high-level planning datasets, such as ShareRobot~\cite{ji2025robobrain} and EO1.5M-QA~\cite{qu2025eo1}.
    These sources explicitly train the model to decompose complex, long-horizon commands into logical sub-task sequences, bridging the critical gap between abstract user intent and low-level motor execution.

\end{enumerate}

\section{UniCraftor: A System for Portable, Extensible, and Affordable Data Collection}

While open-source corpora offer significant scale, they often lack critical supervision signals such as accurate depth, stable camera alignment, and temporally precise interaction events.
For example, Ego4D~\cite{grauman2022ego4d} and its multi-view extension Ego-Exo4D~\cite{grauman2024egoexo4d} provide rich semantic descriptions but lack geometric depth.
Egocentric-10K~\cite{buildaiegocentric10k2025} features 10,000 hours of in-the-wild industrial footage but provides only raw RGB streams without annotation. Furthermore, benchmarks like HD-EPIC~\cite{perrett2025hdepichighlydetailedegocentricvideo} and HOI4D~\cite{liu2022hoi4d} rely on offline calibration to approximate camera poses, and their interaction labels are typically aligned to clip boundaries or a handful of annotated frames.
Such coarse labeling introduces temporal ambiguity regarding the exact moments of object contact and release, which is detrimental to fine-grained policy learning.
To address these deficiencies, we develop a modular, plug-and-play data collection system to curate high-quality, multimodal recordings with synchronized supervision.

\begin{enumerate}[label=\textbf{\arabic*)}]

    \item \textbf{Native Depth Acquisition:} 
    We utilize a head-mounted Intel RealSense D435 for egocentric RGB-D capture.
    By leveraging active infrared stereo to provide raw physical depth, we bypass the artifacts and inconsistencies inherent in learning-based depth estimation, which often degrades under heavy occlusions, rapid ego-motion, or adverse lighting conditions. 
    
    \item \textbf{High-Precision Extrinsics:} 
    To decouple hand-camera motion and establish a consistent world coordinate frame, we compute ground-truth camera poses via five tabletop AprilTags~\cite{olson2011apriltag} using Perspective-n-Point (PnP) algorithms. 
    This approach, rooted in classical robotic hand-eye calibration, offers superior stability against head-motion jitter and sensor noise compared to contemporary learning-based extrinsic predictors~\cite{zhang2025hawor, yu2025dyn}. 
    Furthermore, it enables the seamless integration of exocentric views for flexible and synchronized view captures; additional D435 cameras can be freely repositioned without manual recalibration, provided at least one AprilTag remains within the field of view.

    \item \textbf{Hardware-Synchronized Interaction Events:}
    To resolve the temporal ambiguity in automated labeling where frame downsampling often misses the precise moment of object contact or release, we incorporate a hardware-synchronized foot pedal to record explicit interaction events. 
    This mechanism captures affordance-relevant keyframes with high temporal precision and minimal cognitive overhead for the demonstrator.
    These timestamps are synchronized across all sensor streams, serving as a ground truth for fine-grained semantic descriptions and verifying the temporal accuracy of manipulation cues.

\end{enumerate}

\begin{figure}[h] 
\centering
\includegraphics[width=0.85\linewidth]{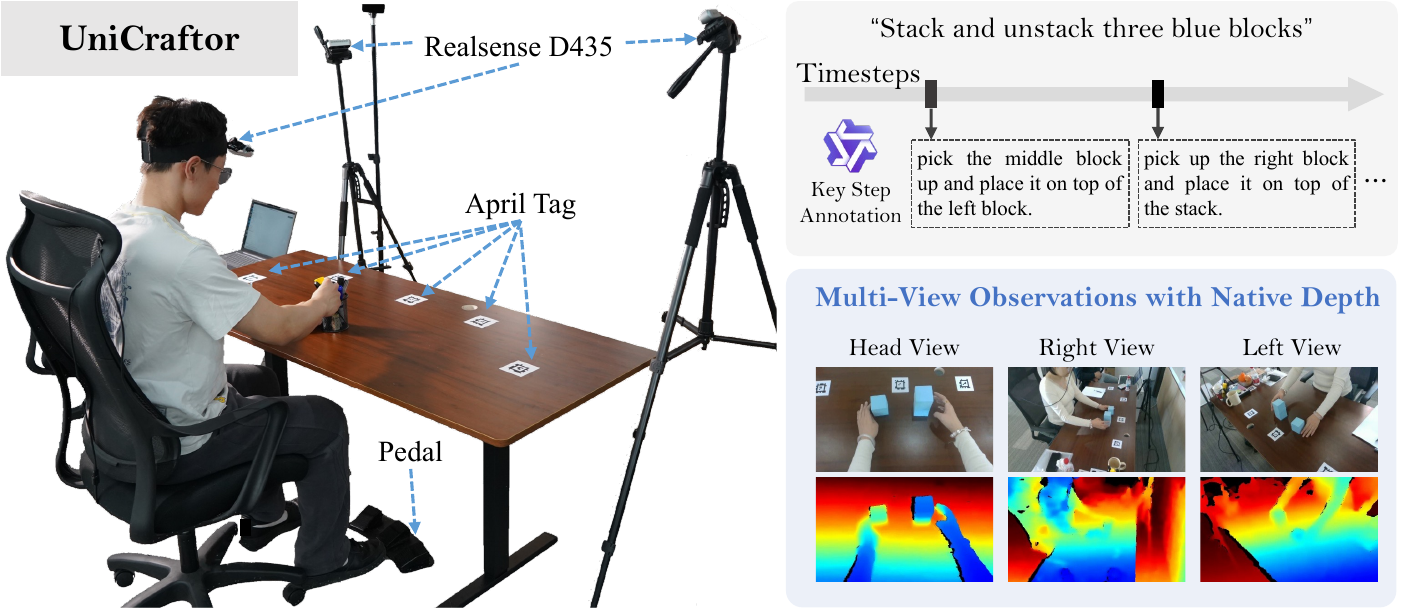}
\caption{An overview of our data collection system UniCraftor.}
\label{fig:data_collect_system}
\end{figure}

Using this system (see Figure~\ref{fig:data_collect_system}), we construct a dataset comprising \textbf{43 tabletop tasks} and \textbf{200+ hours} of multi-modal recordings. 
All sensors are synchronized via a central timestamping device, and the modular architecture permits rapid sensor interchange with planned extensions to tactile sensing and mobile scenarios.
Our post-processing pipeline consists of three stages: 
first, AprilTag regions are tracked and inpainted using Grounded-SAM2~\cite{ren2024grounded} and DiffuEraser~\cite{li2025diffueraser} to ensure visual cleanliness; 
second, hand motions estimated via HaWor~\cite{zhang2025hawor} are refined through multi-view depth integration to enforce cross-view spatial consistency;
finally, fine-grained task descriptions are automatically generated by Qwen2.5-VL~\cite{bai2025qwen2}, conditioned on pedal-triggered event signals and undergo final human verification to ensure groundedness and correctness.

\section{Being-H0.5: A Foundational VLA Unifying Cross-Embodiment Control}

\subsection{Model Architecture}
\begin{figure}[t]
    \centering
    \includegraphics[width=1\linewidth]{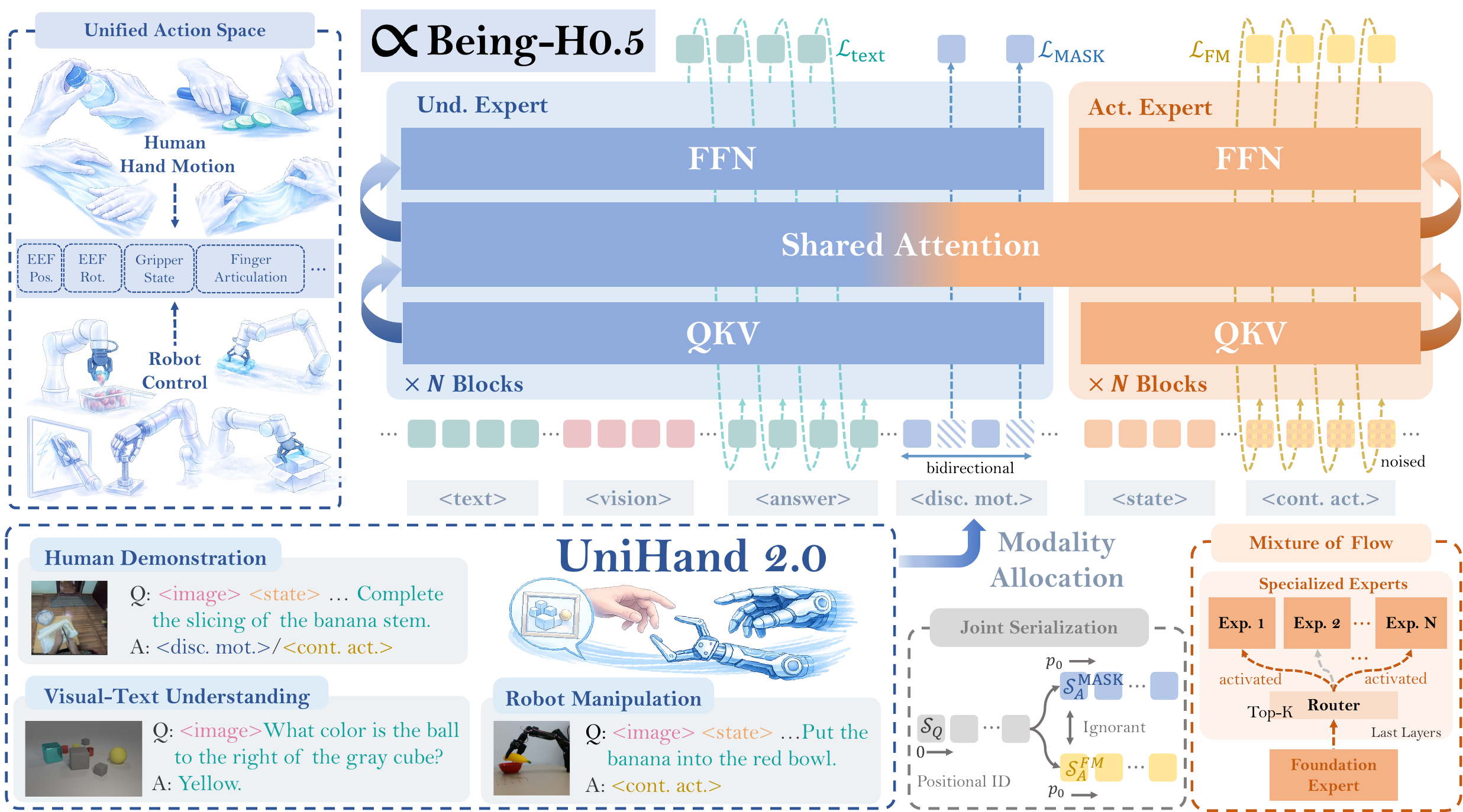}
    \caption{\textbf{Overview of Being-H0.5.} 
    Being-H0.5 is a specialized MoT that disentangles multimodal understanding (\textbf{Und. Expert}) and action generation (\textbf{Act. Expert}) while maintaining coupling through shared attention mechanisms.
    A \textbf{unified state–action space} supports cross-embodiment pre-training by mapping human hand motion and diverse robot controls into semantically aligned slots.
    Our pre-training leverages \textbf{UniHand-2.0} by serializing multimodal data into a unified QA-style format, with each modality allocated to the relevant branch. 
    Finally, a \textbf{Mixture-of-Flow} design scales action capacity by combining shared foundation layers with routed specialized experts for embodiment/task-specific dynamics.}
    \label{fig:model_arch}
\end{figure}

Following the architectural principles of BAGEL~\cite{deng2025bagel}, Being-H0.5 adopts a specialized Mixture-of-Transformers (MoT) architecture~\cite{liang2024mot} designed to disentangle high-level semantic reasoning from low-level motor control.
Within this structure, Being-H0.5 integrates two distinct expert modules:
\textbf{1) Multimodal Understanding Expert} is tasked with interpreting high-dimensional perceptual inputs.
In the context of VLAs, this expert extends beyond traditional image captioning and is engineered for long-horizon planning, generating intermediate subgoals and providing the spatial reasoning necessary to ground the model in complex environments.
\textbf{2) Action Generation Expert} serves as a dedicated policy network responsible for translating high-level plans into precise kinematic execution.

This dual-expert paradigm balances semantic comprehension with continuous control, aligning with the architectural consensus established by recent state-of-the-art models such as $\pi_0$~\cite{black2024pi_0}.
Crucially, while these experts maintain specialized functional roles, they operate within a unified MoT backbone.
Both modules process a common token sequence and leverage shared self-attention mechanisms across each transformer layer.
This design facilitates a seamless information flow, ensuring that action generation is deeply conditioned on visual-semantic context without the overhead of architectural bottlenecks.

Regarding the generation paradigm, we employ a hybrid approach tailored to the modalities of output.
For textual outputs including high-level reasoning and motion description, we utilize the standard next-token-prediction paradigm which leverages the established effectiveness of VLM to produce coherent, logically structured instructional chains, while for discrete hand motion, a masked token prediction criterion is adopted.
For action prediction, we adopt the Rectified Flow~\cite{lipman2022flow, liu2022flow} method, departing from discrete tokenization in favor of continuous actions.
This aligns our policy head with cutting-edge diffusion-based VLA approaches, allowing for the generation of smooth, high-fidelity multimodal distributions over the action space.

The backbone of Being-H0.5 is initialized from InternVL-3.5~\cite{wang2025internvl3.5}, a publicly available VLM featuring a robust decoder-only transformer architecture.
We select InternVL-3.5 for its proven performance on complex visual reasoning benchmarks and accessibility.
It is worth noting that the choice of the VLM backbone is critical, with empirical evidence suggests that the underlying visual features significantly dictate downstream VLA efficacy.
Given that architectural biases can lead to substantial disparities in robotic performance, we intend to systematically benchmark alternative backbones in future exploration.

\subsubsection{Unified State-Action Space}
A primary obstacle to scaling generalist robot learning is the extreme heterogeneity inherent in multi-source embodiment data.
The ``physical gap'' between human hands and robotic end-effectors, combined with variances in kinematics, actuation limits, and control frequencies across diverse robotic platforms, results in a highly fragmented data landscape. 
Furthermore, existing datasets frequently utilize inconsistent nomenclature and disparate physical units for semantically identical actions.
Prior approaches~\cite{nvidia2025gr00t} often circumvent these discrepancies by assigning independent MLP projection layers (i.e., separate encoder/decoder heads) to each specific embodiment to handle dimensional variations.
We posit that this strategy is suboptimal, as it unnecessarily consumes model capacity and fragments the learning of physical commonalities.
In practice, diverse gripper configurations and dexterous hands exhibit high geometric consistency in End-Effector (EEF) trajectories, and many joint configuration possess underlying alignment potential. 
Isolating these through separate MLP heads prevents the model from leveraging shared physical priors, thereby degrading cross-embodiment generalization. 
Additionally, many existing methods tolerate mixed rotation representations (e.g., Euler angles, quaternions), forcing the network to squander computational resources on non-essential formatting discrepancies.
This lack of structure introduces unnecessary noise, destabilizing the training process and severely limiting the model's ability to transfer knowledge from human video data to robotic control.

To bridge these kinematic disparities, we introduce a \textbf{physically interpretable unified state-action space}.
We formalize both state and action as fixed-length, high-dimensional vectors structured by the principle of \textbf{physical semantic alignment} proposed in ~\cite{luo2025beingh0}.
Conceptually, we partition the vector space into semantically isolated subspaces where every dimension corresponds to a grounded physical quantity, such as bimanual EEF poses, joint positions, gripper width, finger articulations, or mobile base velocity and heading commands.
A critical innovation in our unified space is the treatment of \textbf{human hand motion as a generalized embodiment}. 
We map parameters from the MANO hand model directly into this unified space.
Specifically, the global wrist pose of the human hand is aligned with the robotic EEF subspace, while finger articulations are mapped to reserved ``fine-manipulation'' slots.
This architecture ensures that distinct degrees of freedom (DoF) do not conflict, allowing the model to learn a shared, embodiment-agnostic latent representation of manipulation logic.
By utilizing this common ``action language'', Being-H0.5 can be effectively pre-trained on vast libraries of human videos and episodes of different embodiment while remaining directly compatible with downstream robotic execution.

To guarantee transferability across heterogeneous platforms, we enforce strict standardization within each subspace.
For Cartesian control, all end-effector actions are expressed as relative delta displacements in a unified world coordinate frame.
Rotations are uniformly parameterized using Axis-Angle notation to prevent gimbal lock and ensure smooth interpolation on the SE(3) manifold.
For joint-space control, positions are standardized as absolute radian values. 
Notably, we eschew traditional statistical normalization (e.g., scaling to [-1, 1]) in favor of preserving raw physical magnitudes.
We argue that a movement of 1 radian or 10 centimeters carries intrinsic physical implications that normalization obscures.
By applying only outlier filtering to mitigate sensor noise, we force the model to learn the true physical scale of actions, resulting in a policy that is both generalizable and physically grounded across diverse embodiments and environments.

\subsubsection{Mixture of Flow}
While a unified state-action space effectively represents diverse embodiments and prevents representational conflicts during pre-training, the limited capacity of conventional action experts remains a critical bottleneck.
This capacity constraint often leads to performance degradation in VLAs when integrating robots with heterogeneous morphologies --- particularly as action expert parameters are typically far fewer than those in flow-based experts for visual generation~\cite{deng2025bagel, seedream2025seedream}.
Furthermore, such restricted capacity hampers the model’s ability to generalize across a broad spectrum of embodiments and complex downstream tasks.

To overcome these limitations, we introduce \textbf{Mixture of Flow (MoF)}: a scalable architectural framework designed to decouple distinct embodiments and skills while leveraging a shared foundational representation.
Our approach is motivated by two key observations.
First, many embodiments share partial control structures despite significant visual and kinematic differences.
For instance, both Franka and Kuka robots are N-DoF manipulators defined by joint positions and end-effector states, while mobile platforms like wheeled half-humanoids share common velocity and orientation metrics for base control.
Decomposing a unified expert into specialized sub-modules allows the model to master specific state-space intervals, thereby mitigating inter-embodiment interference.
Second, human motor control is inherently modular and general motor primitives are dynamically adapted to specific tasks.
Reflecting these principles, our flow-based action expert is structured into a two-tiered hierarchy:
\textbf{1) Foundation Experts (Shared Dynamics):} 
The initial layers of the action expert consist of standard transformer blocks shared across all inputs.
These layers encode fundamental, transferable motor primitives (e.g., reaching, grasping dynamics, and collision avoidance) that remain invariant across disparate embodiments and tasks.
\textbf{2) Specialized Experts (Embodiment \& Task Routing):}
The upper layers utilize a suite of parallel, specialized experts managed by a learnable gating network, inspired by Mixture-of-Experts (MoE) architectures~\cite{shazeer2017moe}. 
For a given input state and instruction, the router dynamically activates a sparse subset (e.g., Top-K) of experts, facilitating efficient specialization without a linear increase in computational overhead.

This architectural sparsity is fundamental to the efficiency and robustness of Being-H0.5.
It allows the model to synthesize refined foundation primitives into complex, task-specific behaviors without cross-task interference. 
During training, gradients for a specific task update only the relevant expert pathway, thereby preserving the weights of other localized skills.
Furthermore, this design decouples the total parameter count from the active parameter count.
While the unified model hosts an extensive library of skills, only a fraction of its parameters are activated during inference.
This makes Being-H0.5 highly deployable on resource-constrained edge hardware, such as the NVIDIA Orin-NX, allowing the robot to access broad, on-demand capabilities without exceeding memory bandwidth or latency constraints.

\subsection{Pre-Training: Human-Centric Robot Learning}
Our robot learning framework is fundamentally human-centric, utilizing the UniHand-2.0 dataset as its structural cornerstone.
In this paradigm, human behavior is treated as a dense source of physical priors rather than a passive reference.
UniHand-2.0 serves a dual role: its large-scale egocentric human data provides transferable behavioral intent to encode how complex interactions unfold in the open world, while its robot trajectories provide the high-fidelity kinematic supervision essential for low-level motor control.

To bridge the gap between human demonstration and robotic execution, we incorporate general Visual Question Answering (VQA) signals.
This integration injects broad vision-language context and enhances scene understanding, addressing a critical dichotomy in embodied intelligence: the necessity for precise motor control coupled with rich contextual grounding.
By synthesizing these heterogeneous streams, our multi-task objective trains the model to perceive with the nuance of a VLM while executing with the precise action.

\subsubsection{Unified Sequence Modeling}
We implement our learning framework by \textbf{casting all supervision into a unified, multimodal sequence modeling problem}. 
By treating the state-action space as an explicit modality, we define a state vector $\mathbf{s}\in\mathbb{R}^{d}$ and an action vector $\mathbf{a}\in\mathbb{R}^{d}$ shared across all embodiments.
Each embodiment $e$ (e.g., parallel grippers, dexterous hand, or mobile bases) is projected into this unified space via sparse slot assignments:
\begin{equation}
\mathbf{s} = \Phi_e(\mathbf{s}^{(e)}), \quad 
\mathbf{a} = \Phi_e(\mathbf{a}^{(e)})
\end{equation}
where $\mathbf{s}^{(e)}$ and $\mathbf{a}^{(e)}$ represent raw, embodiment-specific signals.
The mapping function $\Phi_{e}$ routes these signals to relevant slots within the global vector, leaving unused slots as zeros.
This allows human hand motion and robot trajectories to co-exist within a single, physically interpretable interface alongside vision and language.
Formally, each training sample is serialized into a token stream $\mathcal{S}$ composed of $K$ modality segments:
\begin{equation}
\mathcal{S} = [ \mathbf{x}_1, \mathbf{x}_2, \dots, \mathbf{x}_K ]
\end{equation}
where each segment $\mathbf{x}_k = \langle m_k, C_k \rangle$ consists of a modality tag $m_k \in \mathcal{M}$ and its corresponding content $C_k$. 
The modality set is defined as:
\begin{equation}
\mathcal{M} = \{ \text{vision, text, state, action} \}
\end{equation}
Each sample instantiates a subset $\mathcal{M}_i \subseteq \mathcal{M}$ based on available supervision, and each non-text modality is encapsulated within two distinct special tokens. 
During training, we adopt \textbf{Physical Instruction Tuning}~\cite{luo2025beingh0} to organize data into a Query-Answer format $[\mathcal{S}_Q ; \mathcal{S}_A]$.
The model is conditioned on the context $\mathcal{S}_Q$ and optimized via a generative loss exclusively on the response $\mathcal{S}_A$.

\subsubsection{Human-Centric Multi-Task Objective}
Within the unified sequence, we define a family of tasks distinguished by their modality organization.
First, \textbf{motion generation} serves as the primary objective for manipulation learning
The model predicts an \texttt{action} chunk conditioned on \texttt{vision}, \texttt{text}, and \texttt{state}. 
This is the main pathway through which large-scale human interaction traces and robot trajectories provide policy-relevant supervision.
To complement this, we introduce \textbf{motion description} and \textbf{continuation} as alignment objectives to strengthen vision-language-motion grounding and temporal coherence.
For motion description, the model predicts \texttt{text} conditioned on \texttt{vision} and interaction traces (\texttt{state}/\texttt{action}), which enforces semantic grounding, aligning high-level linguistic intent with physically grounded outcomes.
For motion continuation, the model predicts future \texttt{action} chunks conditioned on past observations and action history, encouraging the learning of temporally consistent interaction dynamics beyond single-step imitation.

While our pre-training is fundamentally human-centric,  with egocentric human data forming the core of our data recipe, the QA-pair serialization seamlessly absorbs diverse auxiliary supervision.
By instantiating only available modalities, the framework incorporates robot manipulation data for action generation ($\mathcal{S}_A = \{\texttt{action}\}$) and VQA datasets for standard text prediction ($\mathcal{S}_A = \{\texttt{text}\}$).
All tasks share a common backbone and are optimized via a joint loss function:
\begin{equation}
\mathcal{L} =
\lambda_{\text{text}}\mathcal{L}_{\text{text}}
+
\lambda_{\text{act}}\mathcal{L}_{\text{act}}.
\end{equation}
To handle the multimodal nature of $\mathcal{S}_A$, we define token-level index sets $\Omega_{\text{text}}$ and $\Omega_{\text{act}}$ to identify segments of the unified sequence subject to each loss term.
For VQA and motion-description tasks, we apply a standard cross-entropy loss over the target text tokens:
\begin{equation}
\mathcal{L}_{\text{text}}
= -\sum_{i\in\Omega_{\text{text}}}\log p_\theta(y_i \mid \mathcal{S}_{<i}).
\end{equation}
For motion generation and continuation (from both human and robot sources), we supervise the action tokens using a dedicated action loss $\mathcal{L}_{\text{act}}$, the formulation of which---dependent on the specific motion representation---will be detailed in the following section.

\subsubsection{Hybrid Human Motion Representation}
\label{sec:dual_supervision}
Human hand motion is inherently expressive; however, at scale, it inevitably exhibits subtle execution variations and observational noise.
Relying exclusively on continuous supervision can result in brittle learned priors, while purely discrete representations may sacrifice the precision necessary for fine-grained control.
To capture both the high-fidelity requirements of robot control and the stable behavioral priors of human-like movement, we supervise motion using two complementary representations within the same training instance: 
1) continuous action chunks in the unified space and 
2) discrete motion tokens derived via chunk quantization, as introduced in Being-H0~\cite{luo2025beingh0}.

Specifically, given a motion chunk of length $T$, we represent it as a continuous sequence
$\mathbf{A} = [\mathbf{a}_1,\ldots,\mathbf{a}_T] \in \mathbb{R}^{T \times d}$, 
where each $\mathbf{a}_t\in\mathbb{R}^d$ resides in our unified, high-dimensional action space. 
Simultaneously, we leverage a pretrained tokenizer to quantize the motion into a discrete sequence $\mathbf{z} \in \{1,\ldots, |\mathbbm{C}|\}^{T_z}$.
Here, $\mathbbm{C}$ denotes a codebook of size $|\mathbbm{C}|$, and $T_z$ represents the resulting tokenized length.
This discrete channel serves as a robust, language-like abstraction that filters high-frequency execution noise, thereby stabilizing the motion prior across heterogeneous datasets.
The joint objective is formulated as weighted combination of \textbf{Continuous Flow-Matching} and discrete \textbf{Masked Motion Prediction}:
\begin{equation}
\mathcal{L}_{\text{act}}
=
\lambda_1\mathcal{L}_{\text{FM}}
+
\lambda_2\mathcal{L}_{\text{MASK}}.
\end{equation}

\textbf{Continuous Flow-Matching.}
We model the continuous motion distribution using a time-conditioned velocity field $v_\theta(\mathbf{x},t,c)$, where $c$ represents the conditioning visual and textual context.
This field is designed to transport samples from a standard Gaussian distribution $\mathbf{x}_0\sim \mathcal{N}(\mathbf{0},\mathbf{I})$ toward the target data distribution.
For a target action $\mathbf{a}_i$ at temporal index $i$, we define a probability path via linear interpolation: $\mathbf{x}_t = (1-t)\mathbf{x}_0 + t\mathbf{a}_i$ for $t \in [0, 1]$.
The objective $\mathcal{L}_{\text{FM}}$ minimizes the mean squared error between the predicted velocity and the ideal vector field $\mathbf{u}_t(\mathbf{x}_t) = \mathbf{a}_i - \mathbf{x}_0$.
Let $\Omega_{\text{FM}}$ denote the set of indices for continuous action steps within the answer segment $\mathcal{S}_{A}^{\text{FM}}$. The objective can be denoted as:
\begin{equation}
\mathcal{L}_{\text{FM}}
=
\sum_{i\in\Omega_{\text{FM}}}
\left\|
v_\theta(\mathbf{x}_t,t,c)
- (\mathbf{a}_i - \mathbf{x}_0)
\right\|_2^2.
\end{equation}

\textbf{Masked Motion Token Prediction.}
In addition to continuous regression, we supervise the discrete channel to instill a stable, abstraction-level of motion primitives.
Specifically, we augment the model backbone with a dedicated token embedding and a linear prediction head for quantized tokens.
During training, we apply a masking strategy to the token sequence $\mathbf{z}\in\{1, \ldots, |\mathbbm{C}|\}^{T_z}$ by randomly sampling a subset of indices $\Omega_{\text{MASK}}\subseteq\{1,\ldots,T_z\}$ according to a masking ratio $\rho$, replacing the original tokens with a specialized \texttt{[MASK]} token.
The model is tasked to reconstruct the original codebook indices via a cross-entropy loss:
\begin{equation}
\mathcal{L}_{\text{MASK}}
= -\sum_{i\in\Omega_{\text{MASK}}}\log p_\theta(z_i \mid c).
\end{equation}
By predicting these discrete tokens, the model learns the underlying ``grammar'' of hand motion, providing a structural scaffold that supports the continuous flow-matching head.

\paragraph{Joint Serialization.}
To achieve seamless multi-modal integration, we serialize the context and hybrid targets into a unified sequence.
For a given instance, we concatenate a shared context prefix $\mathcal{S}_Q$ with two target segments: $[\mathcal{S}_Q \,;\, \mathcal{S}_A^{\text{FM}} \,;\, \mathcal{S}_A^{\text{MASK}}]$.
Here, $\mathcal{S}_{A}^{\text{FM}}$ contains continuous action targets and $\mathcal{S}_{A}^{\text{MASK}}$ contains discrete motion-token targets, both conditioned on the same context $\mathcal{S}_Q$. 
To prevent trivial information leakage where the model might copy information between the continuous and discrete channels, we enforce both target segments $\mathcal{S}_A^{\text{FM}}$ and $\mathcal{S}_A^{\text{MASK}}$ can attend to the shared context $\mathcal{S}_Q$ but remain mutually invisible.
This is implemented via a modified attention mask with a gating matrix:
\begin{equation}
G \;=\;
\begin{bmatrix}
\textbf{1} & \textbf{0} & \textbf{0}\\
\textbf{1} & \textbf{1} & \textbf{0}\\
\textbf{1} & \textbf{0} & \textbf{1}
\end{bmatrix},
\qquad
\text{for} \ \ [\,\mathcal{S}_Q \,;\, \mathcal{S}_{A}^{\text{FM}} \,;\, \mathcal{S}_{A}^{\text{MASK}}\,].
\end{equation}
The effective attention mask is implemented by applying this gating onto the original mask.
Furthermore, we align the positional indices to ensure that both targets are grounded from the same positional origin. 
Let $r(j)$ be the relative index within a segment, and $p_0=\max_{j\in\mathcal{S}_Q}(j+1)$. 
The positional encoding is defined as:
\begin{equation}
\text{PE}(j)=
\begin{cases}
j, & j \in \mathcal{S}_Q,\\
p_0 + r(j), & j \in \mathcal{S}_A^{\text{FM}},\\
p_0 + r(j), & j \in \mathcal{S}_A^{\text{MASK}}.
\end{cases}
\end{equation}
This alignment allows the model to perceive the continuous and discrete representations as two complementary views of the same temporal event, grounded to an identical contextual origin.

\subsection{Post-Training: Towards Cross-Embodiment Adaptation}
\label{sec:posttrain}

While the UniHand-2.0 dataset provides Being-H0.5 extensive embodiment diversity, yielding stronger \emph{structural priors} than existing VLAs, the \textbf{post-training} phase remains a non-trivial challenge.
The primary objective is to adapt the pre-trained policy to specialized deployment robots, which often possess unique kinematics, constraints, and runtime characteristics, without eroding the general-purpose representations acquired during large-scale pretraining.
In practice, this stage is dominated by a plasticity–stability dilemma: downstream fine-tuning can easily collapse into a narrow specialist, achieving high success within a limited trajectory distribution at the expense of reasoning depth and cross-embodiment transferability.

We identify three coupled sources of brittleness in cross-embodiment deployment:
1) morphological interference from embodiment-dependent action fields competing under shared parameters;
2) context unreliability under distribution shift, where degraded features induce action jitter in flow updates; and
3) real-time temporal mismatch between inference and execution under heterogeneous latency/frequency.
To address them, we introduce two complementary post-training techniques (\textbf{Embodiment-Specific Adaptation} and \textbf{Manifold-Preserving Gating}), as well as a deployment-aware training protocol, \textbf{Universal Async Chunking}.

\subsubsection{Embodiment-Specific Adaptation}
\label{sec:esa}

Despite the powerful vision-language-action representations learned during pre-training, the model must still account for the unique physical dynamics of target hardware.
Each embodiment $e$ induces a different feasible action set governed by kinematic limits, self-collision, and contact constraints.
consequently, the conditional distribution $p(\mathbf{a}\mid H,e)$ becomes increasingly multimodal and less smooth as embodiment complexity increases.
From a flow-matching view, the action expert learns an embodiment-conditioned velocity field $v_{\theta}(\mathbf{a}_t;H,e)$ to match the target field $v^*(\mathbf{a}_t,\mathbf{a}_0)$:
\begin{equation}
\min_{\theta}\ \mathbb{E}_{e}\,\mathbb{E}_{(\mathbf{a}_0,H)\sim \mathcal{D}_e}\,\mathbb{E}_{t}\,\big\| v_{\theta}(\mathbf{a}_t;H,e)-v^*(\mathbf{a}_t,\mathbf{a}_0)\big\|_2^2.
\end{equation}
When optimal fields differ substantially across embodiments, embodiment-specific gradients $\nabla_{\theta}\mathcal{L}_e$ become misaligned (morphological interference), slowing convergence and yielding unstable or ``averaged'' behaviors on complex hardware.

\textbf{Key Design Goal.} 
Our objective is to localize specialization to embodiment-relevant action components while maintaining the stability of the shared VLM backbone.
Notably, we eschew the conventional strategy of assigning a separate projection head to each robot.
Instead, ESA leverages our unified action space: different embodiments activate different index sets (with partial overlap for shared morphological components), and we only update parameters tied to these active indices.

\textbf{Slot-Wise Adapters in a Unified Action Space.}
Let $\mathcal{S}=\{1,\ldots,K\}$ denote the $K$ semantic action slots (\eg, arm joints, hand/gripper joints, base controls), and let $\mathcal{I}_e\subseteq\mathcal{S}$ denote the active slot indices for embodiment $e$.
Embodiments sharing hardware components (\eg, the same arm with different hands) feature overlapping index sets and therefore share a subset of adaptation parameters.
We maintain a slot-wise bank of lightweight adapters
$\mathbf{W}_{\text{ESA}}\in\mathbb{R}^{K\times d_{\text{out}}\times d_{\text{in}}}$,
and during post-training, update only the subset indexed by $\mathcal{I}_e$ on embodiment $e$:
\begin{equation}
\mathbf{W}_{\text{ESA}}^{(e)} \triangleq \{\mathbf{W}_{\text{ESA}}[k] : k\in \mathcal{I}_e\},
\qquad
\Delta \mathbf{W}_{\text{ESA}}[k]=\mathbf{0}\ \ \forall k\notin \mathcal{I}_e.
\end{equation}
Operationally, the ESA update is masked by the slot selector $\mathcal{I}_e$: only adapters tied to active slots contribute gradients, so parameters tied to shared slots can transfer knowledge across embodiments, while non-overlapping slots remain isolated.
The result is a single checkpoint capable of supporting heterogeneous robots without parameter conflict.

\subsubsection{Manifold-Preserving Gating}
\label{sec:mpg}

\begin{figure*}[t]
    \centering
    \includegraphics[width=\textwidth]{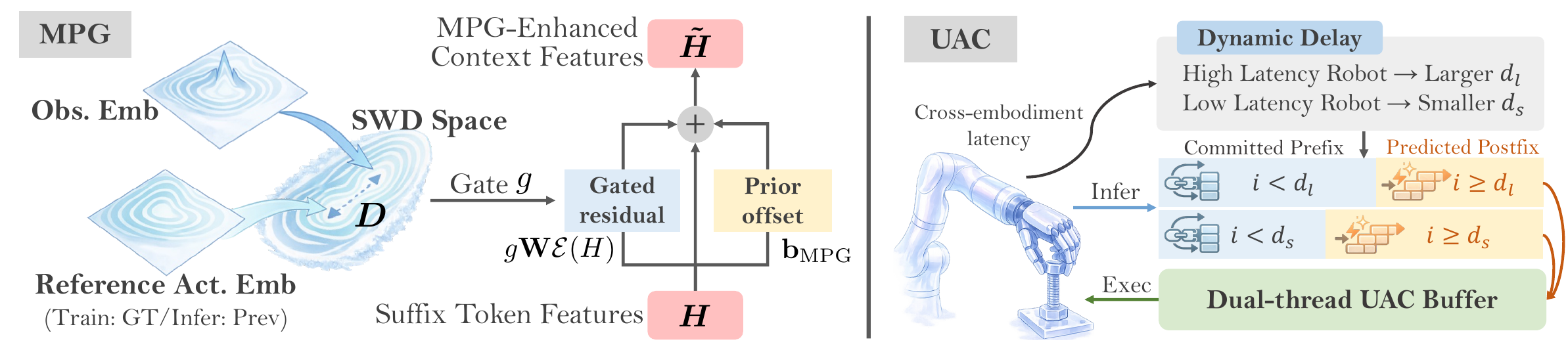}
    \caption{\textbf{MPG and UAC Overview.}
    \textbf{Left (MPG):} 
    We compare observation embeddings with a reference action embedding (Train: ground truth; Inference: previous iterate) in the Sliced-Wasserstein Distance (SWD) space to obtain a discrepancy-guided gate $g$. The gate scales a \emph{feature-conditioned} residual while an \emph{ungated} learned prior offset provides a stable fallback, producing enhanced context features $\tilde H$ for the action expert.
    \textbf{Right (UAC):} 
    Based on embodiment-specific dynamic delay $d$, each predicted action chunk is split into a committed prefix $\mathbf{A}_{<d}$ (already queued/executing) and a predicted postfix $\mathbf{A}_{\ge d}$. 
    A dual-thread buffer enables asynchronous inference/execution across robots with heterogeneous latency budgets.}
    \label{fig:mpg_uac}
\end{figure*}

Flow matching transports noise into the action space conditioned on token-level context features $H$ consumed by the action expert.
In our implementation, $H$ refers to the \emph{suffix token features} formed by concatenating proprioceptive state embeddings and current action-token embeddings (projected to the VLM hidden size), while the language/vision prefix remains unchanged.

\textbf{Motivation.}
A critical but often implicit assumption is that $H$ faithfully captures task-relevant semantics.
Under distribution shift (\eg, lighting changes, viewpoint perturbations, partial occlusions), a policy may blindly regress actions from degraded features, producing unstable behavior.
This issue is amplified in cross-embodiment adaptation: higher-DoF embodiments introduce additional unobserved state and stronger constraints, making the mapping from $H$ to feasible actions less deterministic and more sensitive to small feature errors.
In flow-matching, a denoising update with step size $\Delta t$ is
$\mathbf{a}_{t-\Delta t}=\mathbf{a}_{t}+\Delta t\, v_{\theta}(\mathbf{a}_t;H)$;
thus perturbations in context features (\eg, $H=H^{\star}+\epsilon$) can induce action variance proportional to
$\big\|\partial v_{\theta}/\partial H\big\|^2\,\mathrm{Var}(\epsilon)$, manifesting as action jitter.

Inspired by DiG-Flow~\citep{zhang2025dig}, MPG treats $H$ as a \emph{noisy conditioning signal} and explicitly estimates its reliability.
When $H$ is out-of-distribution for a target embodiment, conditioning the action expert on $H$ can inject spurious signals and force the model to fit inconsistent velocity fields, often resulting in jittery actions.
MPG therefore utilizes a discrepancy-guided gate to \emph{down-weight feature-dependent corrections} while retaining an \emph{ungated} learned prior offset for graceful fallback (see Figure~\ref{fig:mpg_uac}, left).

\textbf{Gated Residual with an Ungated Prior Offset.}
MPG computes a gate $g\in(0,1]$ (large discrepancy $\Rightarrow$ small $g$) and modulates only the \emph{feature-conditioned residual pathway}:
\begin{equation}
\tilde H
= H + \lambda \cdot \mathcal{P}_{\text{MPG}}\!\big(g \cdot \mathcal{E}_{\text{obs}}(H)\big)
= H + \lambda g\, \mathbf{W}_{\text{MPG}}\mathcal{E}_{\text{obs}}(H) + \lambda \mathbf{b}_{\text{MPG}},
\label{eq:mpg_residual}
\end{equation}
where $\mathcal{E}_{\text{obs}}$ projects $H$ to a common embedding space and
$\mathcal{P}_{\text{MPG}}(\cdot)=\mathbf{W}_{\text{MPG}}(\cdot)+\mathbf{b}_{\text{MPG}}$ generates the enhancement residual.
This yields an intuitive fallback:
when $g\approx 1$, MPG trusts the current context and allows strong feature-conditioned adaptation;
when $g\approx 0$, the feature-dependent term is suppressed, and the model falls back to a stable, learned prior offset $H+\lambda\mathbf{b}_{\text{MPG}}$.
Because $\mathbf{b}_{\text{MPG}}$ is \emph{ungated}, it continues to influence $\tilde H$ even in low-confidence regimes, effectively learning a robust default correction for uncertain contexts.

\textbf{Variance Reduction vs.\ Output Gating.}
Unlike conventional output gating ($\tilde{H} = H + \lambda g \mathcal{R}(H)$ in DiG-Flow~\citep{zhang2025dig}), MPG applies the gate before the projection.
Under conditions of gate fluctuations (\eg, due to visual jitter), output gating scales both the projected term and the bias, inducing variance proportional to
$\|\mathbf{W}\mathcal{E}_{\text{obs}}(H) + \mathbf{b}\|_2^2$.
In contrast, MPG scales only the projected term, yielding variance proportional to
$\|\mathbf{W}\mathcal{E}_{\text{obs}}(H)\|_2^2$.
Since the ungated offset remains constant w.r.t.\ $g$, MPG produces smoother trajectories under contextual perturbation.

\textbf{Discrepancy and Gate Computation.}
To assess context reliability, we construct an action prior anchor from noise-free actions.
Let $Z^{\text{nf}}$ denote the noise-free action-token embeddings encoded at zero noise level, summarize by mean-pooling
$\bar Z=\text{MeanPool}(Z^{\text{nf}})$.
Intuitively, $\bar Z$ characterizes the manifold of \emph{plausible} action embeddings under the model's training distribution.
We quantify the feature--action distributional discrepancy in a shared, scale-invariant space using the sliced Wasserstein distance (SWD)~\citep{bonneel2015sliced,kolouri2019generalized}:
\begin{equation}
D(\mu_{\hat H}, \mu_{\hat Z}) \approx \frac{1}{M}\sum_{m=1}^{M}
\left\| \text{sort}(\theta_m^\top \hat H) - \text{sort}(\theta_m^\top \hat Z) \right\|_2^2,
\label{eq:mpg_swd}
\end{equation}
where $\theta_m$ represents random unit directions,
$\hat H=\text{LN}(\mathcal{E}_{\text{obs}}(H))$,
and $\hat Z=\text{LN}(\mathcal{E}_{\text{act}}(\bar Z))$.
The $\bar Z$ is broadcast along the observation sequence length when computing $D$.
The discrepancy yields a temperature-scaled reliability gate:
\begin{equation}
g = \exp(-D / \tau) \in (0, 1].
\label{eq:mpg_gate}
\end{equation}
In practice, we apply a stop-gradient operation to the gate, $g^{\text{sg}}=\text{stopgrad}(g)$, to prevent degenerate solutions where the model might learn to manipulate $g$ rather than improving underlying feature quality.
With this operation, $g$ behaves as a pure scaling factor on the enhancement pathway.
Ignoring backpropagation through $g$, the Jacobian is approximately:
\begin{equation}
\frac{\partial \tilde H}{\partial H} \approx \mathbf{I} + \lambda g\, \mathbf{W}_{\text{MPG}}\, \frac{\partial \mathcal{E}_{\text{obs}}(H)}{\partial H},
\end{equation}
This formulation ensures that the feature-dependent correction becomes increasingly insensitive when the context is unreliable (small $g$), providing a robust safeguard against out-of-distribution observations.

\subsubsection{Universal Async Chunking}
\label{sec:uac}

\textbf{Motivation.}
Deploying action-chunking policies on physical hardware introduces a fundamental temporal mismatch:
while the model computes a subsequent action chunk, the robot must continue executing previously committed trajectories.
In simulated environments, this latency gap is often negligible.
However, on real-world robot, inference latency directly induces control discontinuities and catastrophic task failures.

Training-Time RTC~\citep{black2025training} mitigates this by simulating inference delay during the training phase:
prefix actions are assigned a clean timestep ($t{=}1$), while postfix actions receive stochastic noisy timesteps ($t{<}1$), with the loss computed exclusively on the postfix.
Nevertheless, RTC is typically instantiated for a single platform.
This remains a significant limitation for a cross-embodiment VLA where diverse robotic platforms operate under heterogeneous control periods and distinct latency budgets.

UAC extends RTC by rendering the delay \emph{embodiment-aware}.
For an embodiment $e$ characterized by a control period $\Delta t^{(e)}$ and an expected inference latency budget $L^{(e)}$, the effective delay in control steps is scaled as $\lceil L^{(e)}/\Delta t^{(e)}\rceil$.
UAC utilizes an embodiment-specific delay distribution and maximum delay threshold, and is trained jointly with ESA, enabling a single checkpoint to accommodate diverse runtime characteristics (see Figure~\ref{fig:mpg_uac}, right).

\textbf{Embodiment-Aware Delay Sampling and Objective.}
We sample a delay
\begin{equation}
d \sim \pi^{(e)}(d), \quad d \in \{0, 1, \ldots, d_{\max}^{(e)}-1\},
\label{eq:uac_delay}
\end{equation}
where both the distribution $\pi^{(e)}$ and maximum delay $d_{\max}^{(e)}$ are configured per embodiment based on its specific control frequency and expected inference latency.
Following RTC, we partition each action chunk into a committed prefix $\mathbf{A}_{<d}$ and a predicted postfix $\mathbf{A}_{\geq d}$, assigning per-token timesteps as follows:
\begin{equation}
t_i = \mathbbm{1}[i < d] + \mathbbm{1}[i \geq d] \cdot t_{\text{base}},
\quad
t_{\text{base}} \sim p(t).
\label{eq:uac_timesteps}
\end{equation}
The training objective is formulated to exclude prefix positions:
\begin{equation}
\mathcal{L}_{\text{UAC}} = \sum_{i \geq d} \left\| \hat{v}_i - v_i^* \right\|_2^2.
\label{eq:uac_loss}
\end{equation}
Combined with embodiment-specific adaptation, UAC trains a unified model to generate smooth continuations across disparate robots, each with its unique delay characteristics, while remaining compatible with real-time asynchronous deployment via a dual-thread buffer, as detailed in Section \ref{sec:dual-thread-deploy}.

\section{Infrastructure: Real-Time Cross-Embodiment Deployment}
\label{sec:infra}

Bridging the gap between high-capacity VLA models and real-world robotic control requires a deployment pipeline that prioritizes \emph{temporal consistency}, \emph{robustness to distribution shift}, and \emph{low-latency execution}.
In this work, we design an inference infrastructure that translates Being-H0.5's cross-embodiment and asynchronous capabilities into robust real-time control.
Concretely, the infrastructure consists of three tightly coupled components:
(i) \textbf{rectified-flow action inference} with cached prefix features for efficiency;
(ii) \textbf{manifold-preserving refinement} powered by MPG to stabilize inference under shift; and
(iii) a \textbf{universal async chunking protocol} with a dual-thread ring-buffer architecture for continuous control across heterogeneous platforms.
We refer the reader to Figure~\ref{fig:mpg_uac} for recalling the MPG/UAC mechanisms.

\subsection{Manifold-Preserving Refinement}
\label{sec:infra_mpg_refine}

\textbf{Rectified-Flow Denoising.}
Action generation follows an iterative denoising process based on rectified flow.
Starting from Gaussian noise $\mathbf{a}^{(0)} \sim \mathcal{N}(0, \mathbf{I})$, we integrate the learned velocity field $v_\theta$ using Euler steps:
\begin{equation}
\mathbf{a}^{(k+1)} = \mathbf{a}^{(k)} + \Delta t \cdot v_\theta(\mathbf{a}^{(k)}, t^{(k)} \mid H), 
\qquad 
\Delta t = 1/K,\ \ t^{(k)} = k/K,
\label{eq:infra_flow_euler}
\end{equation}
where $K$ is the number of denoising steps and $H$ denotes the token-level context features consumed by the action expert.
As described in Section~\ref{sec:mpg}, $H$ consists of a static vision-language prefix and a dynamic suffix (state tokens concatenated with the current action-token embeddings).
During denoising, only the \emph{action-token embeddings} evolve with $\mathbf{a}^{(k)}$, so the vision-language prefix can be cached (kv-cache) for efficiency.

In practice, we keep $K<10$ to balance action fidelity and latency.
This low-step regime is precisely where MPG becomes valuable: by providing a stable fallback under low confidence, MPG can reduce sensitivity to noisy context and allow strong performance with fewer denoising steps (Figure~\ref{fig:mpg_uac}, left).

\paragraph{Two-Stage Refinement with MPG.}
When MPG is enabled at inference time, we use a lightweight two-stage strategy:
(1) a baseline pass produces a reference prediction, and
(2) a small number of refinement rounds use the previous prediction as a proxy anchor to compute discrepancy and gate feature enhancement.

\textbf{(1) Stage 1.} \textit{Baseline prediction (w/o MPG).}
We first run rectified-flow denoising with the raw context features $H$ to obtain an initial prediction $\mathbf{a}^{(0)}_{\text{pred}}$.
This baseline may be suboptimal under a distribution shift, but it provides a concrete reference for constructing the MPG action anchor.

\textbf{(2) Stage 2.} \textit{Iterative refinement (w/ MPG).}
We then perform $N_{\text{ref}}$ refinement rounds ($N_{\text{ref}}\!\in\!\{1,2,3\}$ in practice).
At refinement round $n$, we form a noise-free reference action embedding from the previous prediction:
\begin{equation}
\bar Z^{(n-1)} = \text{MeanPool}\!\left(\text{Enc}(\mathbf{a}^{(n-1)}_{\text{pred}},\,\sigma=0)\right),
\qquad
\mathbf{a}^{(n)}_{\text{pred}} = \text{FlowDenoise}\!\left(H;\bar Z^{(n-1)}\right),
\label{eq:infra_refine}
\end{equation}
where $\sigma$ denotes the action noise level and $\sigma=0$ is noise-free.
Here, MPG is applied \emph{inside the denoising loop} at each Euler step, operating on the current suffix features and using $\bar Z^{(n-1)}$ as the reference action anchor for discrepancy computation (Section~\ref{sec:mpg}).
Intuitively, this creates a feedback loop: better actions $\Rightarrow$ better anchors $\Rightarrow$ more accurate gates $\Rightarrow$ more robust features.

\paragraph{Why MPG Helps in the Low-Step Regime.}
MPG's input-gating formulation (Section~\ref{sec:mpg}) is particularly compatible with small $K$.
When the gate $g$ is low (high discrepancy), MPG suppresses feature-dependent correction and falls back to the ungated prior offset, effectively shifting to
$\tilde H \approx H + \lambda \mathbf{b}_{\text{MPG}}$.
This stabilizes the velocity field prediction $v_\theta(\cdot \mid \tilde H)$ under uncertainty, reducing the ``distance'' the flow needs to traverse from noise to a feasible action mode.
Empirically, performance typically saturates within 2--3 refinement rounds even with $K<10$.

\subsection{Universal Async Chunking Protocol}
\label{sec:infra_uac}

\textbf{Motivation.}
Real robots execute actions continuously while inference happens intermittently.
With chunked action generation, this induces a temporal mismatch: while the model computes a new chunk, the robot continues executing previously committed actions.
If the policy ignores this mismatch, inference latency causes trajectory discontinuities and control stuttering.

Training-Time RTC~\citep{black2025training} addresses this mismatch in \emph{training} by simulating inference delay and computing loss only on the postfix.
Here we formalize a \emph{deployment protocol} that makes this principle explicit and cross-embodiment: we commit a prefix based on latency, lock it throughout denoising, and stitch only the postfix back into the execution buffer (Figure~\ref{fig:mpg_uac}, right).

\paragraph{Latency Commitment.}
Before each inference cycle, we declare a delay $d$ that upper-bounds the number of control steps that will be executed before the new chunk is ready:
\begin{equation}
d \ \ge\  \left\lceil t_{\text{inference}} / t_{\text{control}}^{(e)} \right\rceil + \epsilon_{\text{safety}},
\label{eq:uac_commit}
\end{equation}
where $t_{\text{control}}^{(e)}$ is the embodiment-specific control period and $\epsilon_{\text{safety}}$ is a small margin to absorb timing jitter.
A key property is that \emph{overestimation is safe while underestimation breaks continuity}.
Overestimating $d$ simply increases the locked prefix length; underestimating $d$ means the robot executes actions that the model did not condition on, causing discontinuity.

For cross-embodiment deployment, we configure $d$ per platform using:
(1) \textbf{control period} $t_{\text{control}}^{(e)}$ (higher frequency $\Rightarrow$ more steps per second);
(2) \textbf{expected inference latency} $t_{\text{inference}}$ (model/hardware dependent); and
(3) \textbf{safety margin} $\epsilon_{\text{safety}}$ (platform-specific tolerance).

\paragraph{Hard Prefix Locking (denoising-time constraint).}
During denoising, we enforce prefix consistency for all Euler steps:
\begin{equation}
\mathbf{a}^{(k)}_i \leftarrow
\begin{cases}
\mathcal{B}_i, & i < d, \\
\mathbf{a}^{(k)}_i, & i \ge d,
\end{cases}
\qquad \forall k \in \{0,\ldots,K\},
\label{eq:uac_lock}
\end{equation}
where $\mathcal{B}$ is the execution buffer containing the most recently committed actions.
This prevents the flow from ``editing'' actions that are already being executed (or will be executed before inference finishes).

\paragraph{Hard Stitching (buffer update rule).}
After denoising, we discard the prefix $\mathbf{a}_{<d}$ and write only the postfix $\mathbf{a}_{\ge d}$ into the buffer:
\begin{equation}
\mathbf{a}_{\text{exec}}^{(n)} = \mathbf{a}_{<d}^{(n-1)} \ \oplus\  \mathbf{a}_{\ge d}^{(n)},
\label{eq:uac_stitch}
\end{equation}
which guarantees continuity provided the latency commitment~\eqref{eq:uac_commit} is honored.

\subsection{Dual-Thread Deployment Architecture}
\label{sec:dual-thread-deploy}

To realize UAC in real time, we decouple inference and control into separate threads communicating through a shared ring buffer $\mathcal{B}$ (Figure~\ref{fig:mpg_uac}, right):

\begin{itemize}[leftmargin=1.5em, itemsep=2pt]
    \item \textbf{Control thread (consumer).} Runs at fixed frequency $1/t_{\text{control}}^{(e)}$, popping actions from $\mathcal{B}$ and sending them to the robot.
    If the buffer runs low, it holds the last action or executes a platform-specific safe fallback to avoid discontinuities.
    \item \textbf{Inference thread (producer).} Runs asynchronously: it fetches observations, performs rectified-flow denoising (optionally with MPG refinement, Section~\ref{sec:infra_mpg_refine}), and writes \emph{only} postfix actions into $\mathcal{B}$ starting at offset $d$.
\end{itemize}

We size the ring buffer at least $2\times$ the chunk length to reduce underflow risk under typical latency jitter.
A lightweight mutex protects concurrent access during brief read/write operations, ensuring thread safety without introducing meaningful blocking.

\paragraph{Benefits.}
This architecture provides:
(1) \textbf{latency absorption}: inference jitter does not translate to control jitter;
(2) \textbf{graceful degradation}: buffer underflow triggers safe fallback rather than undefined behavior; and
(3) \textbf{cross-embodiment compatibility}: the same protocol supports robots with different control frequencies and compute budgets by adjusting $(d,\mathcal{B})$ parameters and latency margins.

Together, these infrastructure components enable a single generalist Being-H0.5 checkpoint to achieve real-time performance across diverse platforms (e.g., 10\,Hz tabletop manipulators to 50\,Hz humanoids) under both cloud computing (high network latency) and edge deployment (limited computing performance, \eg, NVIDIA Orin NX), while maintaining smooth and continuous control.

\section{Experiment}

\begin{figure}[t]
\centering
\includegraphics[width=1\linewidth]{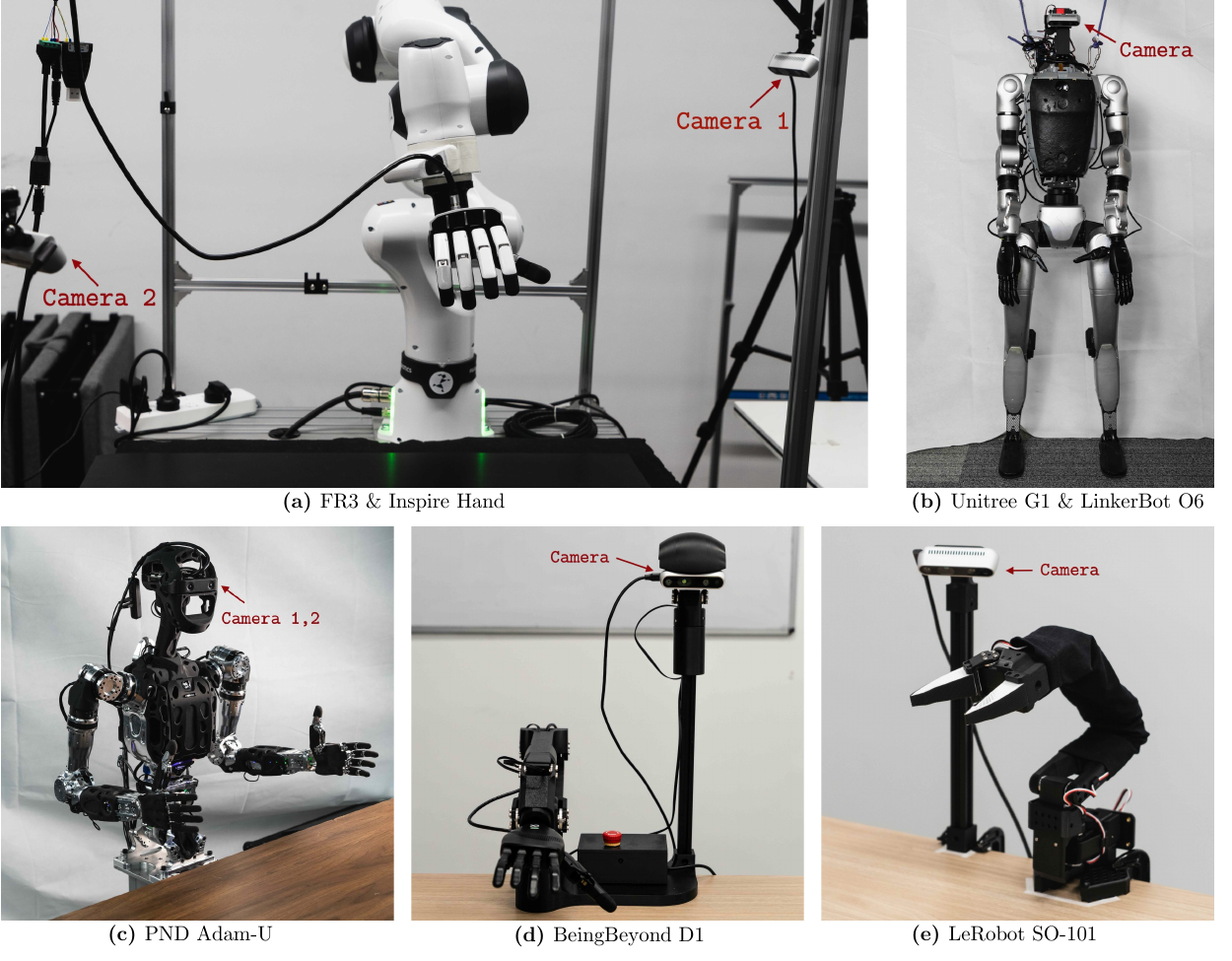}
\caption{\textbf{Real-robot embodiments.} 
We evaluate on five robotic platforms spanning upper-body humanoids, single-arm dexterous manipulation, and lightweight parallel grippers.}
\label{fig:exp_embodiments}
\end{figure}

\subsection{Comparison across Real Robots}
\label{sec:real_robot}

We evaluate Being-H0.5 across five real-world embodiments characterized by substantially different morphologies, sensing setups, and manipulation profiles. 
A total of 10 tasks are designed to rigorously assess four core objectives: 
1) analyzing spatial, long-horizon, bimanual, and generalization capabilities;
2) validate that \textbf{a single checkpoint} can be deployed across heterogeneous embodiments (generalist);
3) quantify the performance gap relative to embodiment-specific optimization (specialist); 
and 4) isolating the architectural and data components critical for real-time robustness (ablation). 

\subsubsection{Experimental Setup}

\begin{table}[t]
\centering
\caption{\textbf{Hardware specifications of heterogeneous real-robot embodiments.}}
\label{tab:real_embodiments_updated}
\resizebox{\linewidth}{!}{
\small
\setlength{\tabcolsep}{6pt}
\renewcommand{\arraystretch}{1.2}
\begin{tabular}{c c c c c c c}
\toprule
\multicolumn{3}{c}{\textbf{Embodiment}} &
\multicolumn{2}{c}{\textbf{Structure}} &
\multicolumn{2}{c}{\textbf{Vision}} \\
\cmidrule(lr){1-3}\cmidrule(lr){4-5}\cmidrule(lr){6-7}
\textbf{Group} & \textbf{Name} & \textbf{DoF} &
\textbf{Joints} & \textbf{Hand} &
\textbf{Camera} & \textbf{Configuration} \\
\midrule

\multirow{2}{*}{Upper-body Humanoid}
& PND Adam-U & 31
& Bimanual+Head+Waist & Dexterous (6DoF)
& ZED Mini& Movable (ego), dual-camera \\

& Unitree G1 + LinkerBot O6 & 26
& Bimanual & Dexterous (6DoF)
& D435 & Fixed (ego) \\

\midrule
\multirow{2}{*}{Single arm + Dexterous hand}
& FR3 + Inspire Hand & 13
& Single-arm & Dexterous (6DoF)
& 2$\times$D435 & Fixed (3rd-person) \\

& BeingBeyond D1 & 14
& Single-arm & Dexterous (6DoF)
& D435 & Movable (ego) \\

\midrule
Single arm + Gripper
& LeRobot SO-101 & 6
& Single-arm & Gripper
& D435 & Fixed (3rd-person) \\

\bottomrule
\end{tabular}
}
\end{table}

\begin{table*}[t]
\centering
\caption{\textbf{Real-robot task suite and instructions.}
Tasks are categorized by their requisite manipulation capabilities.}
\label{tab:real_tasks_prompts}
\resizebox{\textwidth}{!}{
\small
\setlength{\tabcolsep}{6pt}
\renewcommand{\arraystretch}{1.15}
\begin{tabular}{l l l p{10cm}}
\toprule
\textbf{Category} & \textbf{Task} & \textbf{Embodiment} & \textbf{Instruction} \\
\midrule

\multirow{4}{*}{Spatial}
& Arrange Flowers
& Adam-U
& ``Arrange the flowers neatly into the vase.'' \\

& Water Plant
& FR3+Inspire
& ``Use your hand to hold the sprinkling can and water the plants'' \\

& Stack Bowls
& FR3+Inspire
& ``Hold the edge of the bowls and stack them on the plate one by one.'' \\

& Stack Blocks
& D1
& ``Stack the blocks following the color order: \texttt{<ORDER>}.'' \\

\midrule
\multirow{2}{*}{Long-horizon}
& Drawer: Open $\rightarrow$ Place $\rightarrow$ Close
& FR3+Inspire
& ``Open the drawer, put \texttt{<OBJECTS>} inside, then close the drawer.'' \\

& Package: Flip $\rightarrow$ Scan
& G1+O6
& ``Flip the package then scan the barcode with the scanner.'' \\

\midrule
\multirow{2}{*}{Bimanual}
& Hand-over $\rightarrow$ Box
& Adam-U
& ``Use both hands to pick and pass \texttt{<OBJECTS>} into the box.'' \\

& Pack two products $\rightarrow$ Close lid
& G1+O6
& ``Use both hands to put \texttt{<OBJECTS>} into the box, then close the lid.'' \\

\midrule
\multirow{3}{*}{Generalization}
& Wipe Whiteboard
& FR3+Inspire
& ``Use the wiper to clean any marked area of the whiteboard.'' \\

& Clear Table
& Adam-U
& ``Clear the table: put all objects into the box.'' \\

& Clear Table
& SO-101
& ``Clear the table: put all objects into the plate.'' \\

\bottomrule
\end{tabular}
}
\vspace{0.25em}

\end{table*}

The hardware configurations for our experiments are illustrated in Figure~\ref{fig:exp_embodiments}, with evaluated embodiments and sensory setups summarized in Table~\ref{tab:real_embodiments_updated}.
The comprehensive real-robot task suite is detailed in Table~\ref{tab:real_tasks_prompts}.
For each task, we collect 30 to 60 minutes of real-robot demonstration data.
We evaluate Being-H0.5 against the following methods under the same evaluation infrastructure:

\begin{itemize}
    \item \textbf{Being-H0.5-specialist.} Starting from our pre-trained backbone, we conduct embodiment-specific post-training, where the model is fine-tuned independently for each robotic platform using the full task suite collected for that specific embodiment.
    \item \textbf{Being-H0.5-generalist.} A single checkpoint jointly post-trained on all embodiments and all tasks, enabling deployment across the five platforms.
    \item \textbf{Being-H0.5-scratch}: A baseline without UniHand-2.0 pre-training (specialist \& generalist). Utilizing the same MoT architecture and post-training pipeline as Being-H0.5, but initialized without human-centric robot learning via UniHand-2.0.
    We report both specialist and generalist variants to isolate the contribution of large-scale human-robot pre-training to cross-embodiment transfer.
    \item $\boldsymbol{\pi_{0.5}}$. A competitive open-source VLA baseline fine-tuned on the same per-embodiment data. 
    Notably, due to its embodiment-specific action interface (requiring zero-padding to a fixed dim), $\pi_{0.5}$ does not natively support the cross-embodiment generalist setting; it's therefore evaluated exclusively in the specialist regime.
\end{itemize}

\paragraph{Evaluation Protocol.}

\begin{figure}[t]
\centering
\includegraphics[width=\linewidth]{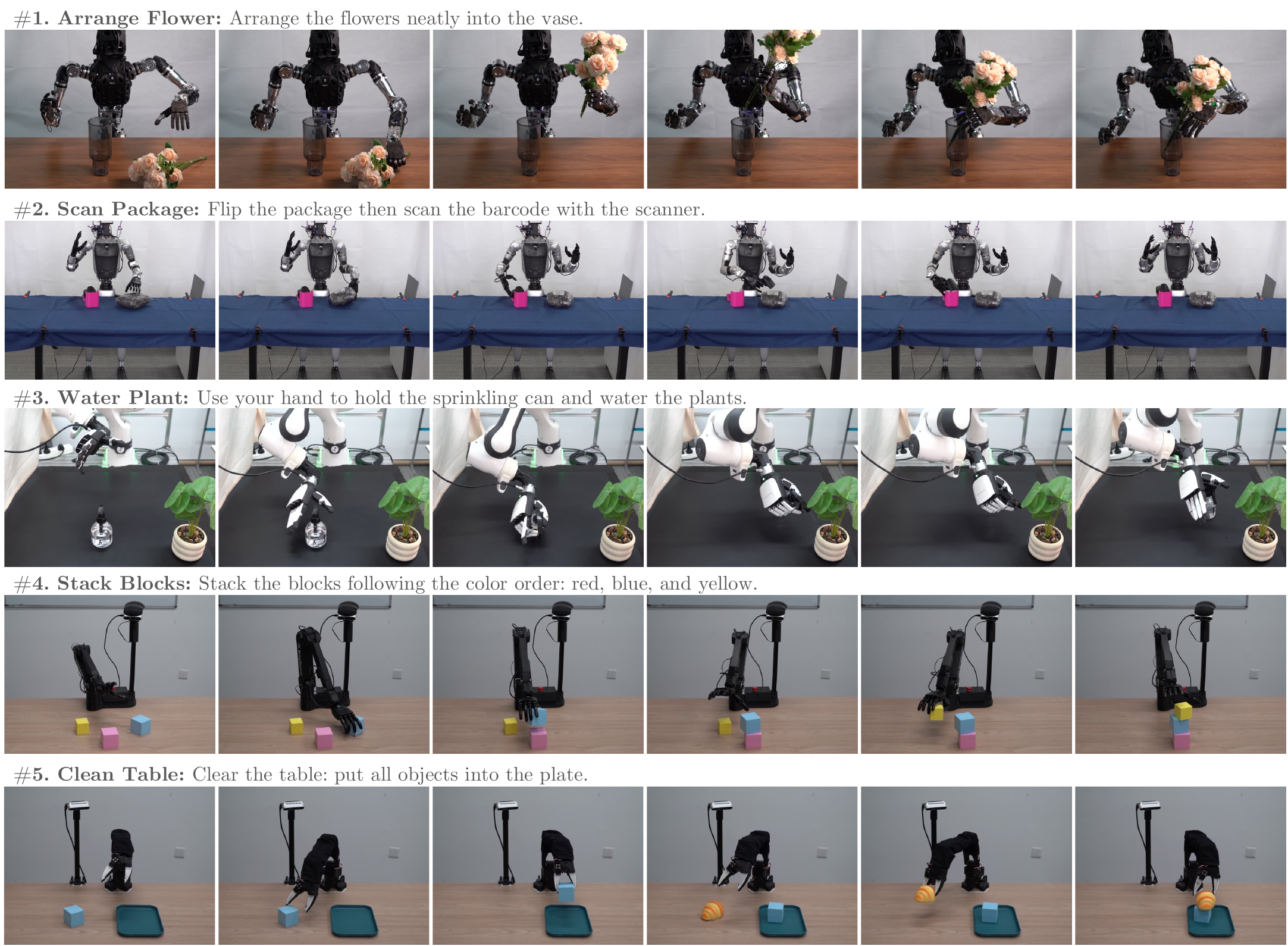}
\caption{\textbf{Example rollouts of representative real-robot task suites.} (Only show one task for each embodiment. The full rollouts of all tasks are shown in Appendix \ref{sec:full_rollouts}).}
\label{fig:exp_traj_main}
\end{figure}

A primary risk in real-robot evaluation is operator bias, which can significantly compromise the fairness of comparisons.
To avoid such bias, we implement a \textbf{blind evaluation protocol} via a unified black-box inference server. The core of this protocol is the \textbf{encapsulation of all candidate policy endpoints into a unified black-box inference server}. 
This server presents a single, standardized interface, ensuring that the evaluation infrastructure is identical for every model. 
Before each evaluation session, we pre-define a comprehensive set of $N$ distinct layouts for each task, systematically varying object positions, orientations, and scene layouts to cover the expected operational distribution and challenge edge cases. 
During evaluation, each trials is executed through the following automated and blinded pipeline: 
\begin{itemize}
    \item \textit{Configuration Sampling.} The system randomly samples one of the $N$ pre-defined configurations and resets the scene accordingly.
    \item \textit{Policy Sampling.} The system randomly selects a policy from the candidate pool. Critically, the policy's identity completely hidden from the human operator. 
    \item \textit{Blinded Execution and Scoring.} The operator will be unaware of which model is being evaluated. Upon task completion or timeout, the operator records a binary success/failure outcome based solely on the pre-defined, objective success criteria.
\end{itemize}

The above process repeats until stopping condition is met: each policy must be evaluated for exactly $K$ trials (default $K{=}20$) under each preset configuration. This protocol guarantees three key properties for statistical validity: 1) all policies encounter identical environmental conditions; 2) the human operator \textbf{cannot consciously or unconsciously favor any particular model}, and 3) the evaluation order is randomized to avoid temporal confounds (\eg, robot wear, lighting drift).

\subsubsection{Result Analysis}
\label{sec:real_robot_results}

\begin{figure}[t]
\centering
\includegraphics[width=.85\linewidth]{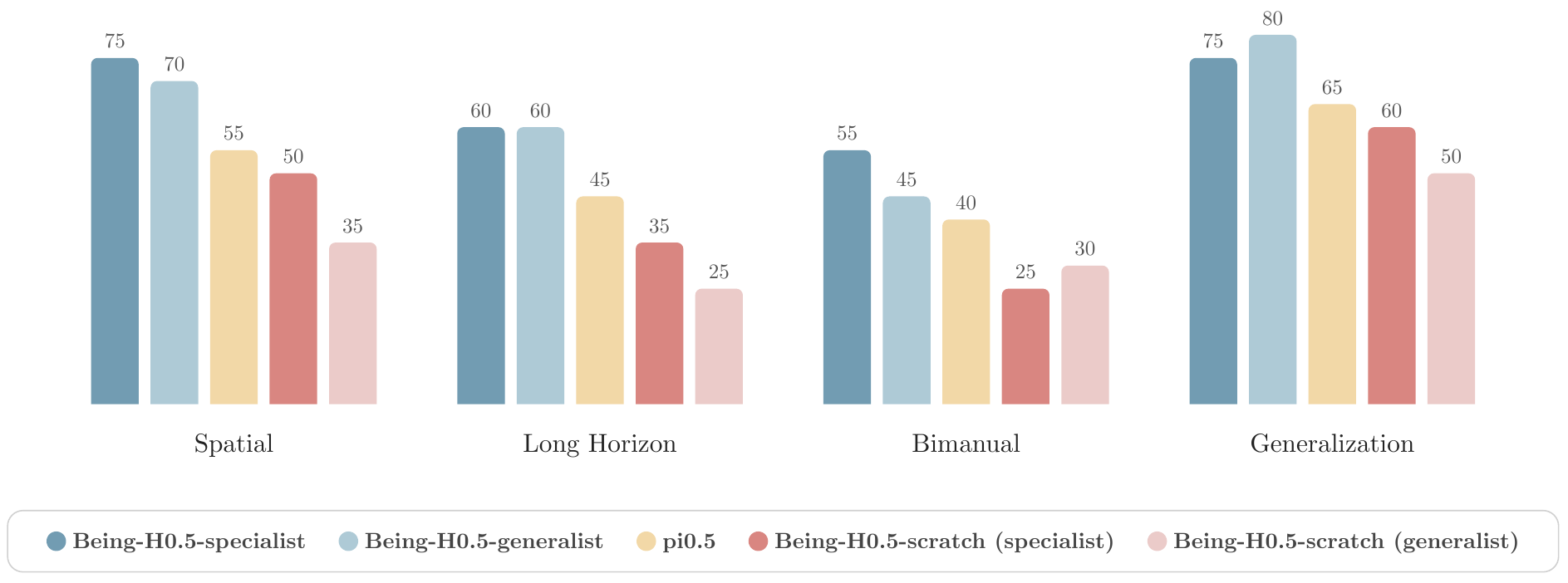}
\caption{\textbf{Real-robot success rates (\%) on four task suites.} We compare Being-H0.5 specialist/generalist, scratch ablations (both specialist and generalist), and $\pi_{0.5}$.}
\label{fig:exp_real_bars_main}
\end{figure}

Figure~\ref{fig:exp_traj_main} shows the example rollouts of the experiments. Figure~\ref{fig:exp_real_bars_main} summarizes category-level success on real robots. We analyze the experimental results in detail in the following section.

\paragraph{Specialist is strongest, but generalist stays surprisingly close.}
Being-H0.5-specialist performs best on most categories, as expected from embodiment-specific adaptation.
Notably, Being-H0.5-generalist is only marginally behind on spatial, long-horizon, and bimanual categories.
In several overlap-heavy settings, the generalist can even be \emph{better} than the specialist.
Empirically, joint training increases exposure to shared sub-skills (approach--grasp--place, lid/handle interactions, planar wiping/clearing),
which improves robustness to nuisance factors (layout shifts, partial occlusions, camera drift) without overfitting to a single embodiment.

\paragraph{Clear margin over $\pi_{0.5}$, especially on long-horizon and bimanual tasks.}
Both Being-H0.5 variants significantly outperform $\pi_{0.5}$.
The gap is most pronounced for long-horizon and bimanual categories, where small perception/control mismatches compound.
A key difference is that Being-H0.5 is trained and deployed in a unified state--action space with an explicit cross-embodiment design,
whereas $\pi_{0.5}$ in our setting remains bound to embodiment-specific action interfaces and thus cannot exploit joint training.

\paragraph{UniHand-2.0 pretraining is crucial for a strong generalist.}
The UniHand-2.0 ablation remains non-trivial in the \emph{specialist} regime (still able to learn from real-robot finetuning),
highlighting that our architecture and post-training pipeline provide a strong inductive bias.
However, the \emph{generalist} ablation degrades substantially and fails to match its specialist counterpart.
This contrast indicates that large-scale cross-embodiment human-centric pretraining is not just ``extra data'': it supplies broad manipulation priors that stabilize optimization under heterogeneous embodiment mixtures and enables positive transfer across platforms.

\paragraph{Embodiment-Level Zero-Shot Transfer.}
Beyond in-distribution evaluation, we observe a notable embodiment-level zero-shot transfer phenomenon.
Specifically, after cross-embodiment post-training, a single \textbf{Being-H0.5-generalist} checkpoint can be deployed on \textbf{Adam-U} to solve \emph{previously unseen task--embodiment pairs}---\ie, tasks for which we collect \textbf{no Adam-U demonstrations at all}.
Concretely, in environments resembling tasks demonstrated only on other embodiments, Adam-U achieves a \emph{non-zero} success rate and consistently exhibits task-consistent multi-step execution.
Examples include: (i) \textit{flip-and-scan} behaviors reminiscent of the G1 package pipeline (reorienting the parcel to expose the label region and bringing the scanner toward it),
(ii) \textit{drawer interaction} behaviors aligned with the FR3 long-horizon task (reaching for handle-like affordances, pulling to open, and attempting to place objects before closing),
and (iii) \textit{stacking} behaviors resembling the D1 block task (sequential pick-and-place with an explicit stacking intent).
While the zero-shot success rate remains lower than in-distribution performance and the motion precision is less reliable, the policy demonstrates a clear ability to \emph{instantiate task structure learned on other morphologies} under a new kinematic chain.

We attribute this capability to two design choices emphasized throughout the paper:
(1) a \textbf{unified action space} that encourages transferable motor primitives rather than embodiment-specific action conventions, and
(2) \textbf{cross-embodiment joint post-training} that exposes the model to diverse realizations of the same underlying task abstractions (objects, goals, and subgoal compositions).
To our knowledge, this is the first empirical evidence that a VLA policy can achieve embodiment-level \textit{zero-shot} task completion (albeit at low success rates) in real-robot deployment.
We believe this result sheds light on a promising scaling direction for VLAs.
In particular, it suggests that increasing the diversity and coverage of post-training data, together with a unified action interface, can pave the way for cross-embodiment policies that generalize compositionally beyond the limited task--embodiment pairs and progressively acquire more transferable, emergent intelligence.

\subsection{Comparison on Simulation Benchmarks}

\begin{table}[h]
\centering
\caption{\textbf{Success rates (\%) on the LIBERO benchmark.} Being-H0.5 is evaluated against a wide range of state-of-the-art VLA models. We report the mean success rate over 50 evaluation episodes per task. Being-H0.5 achieves leading performance on the overall average, with a significant advantage in \texttt{LIBERO-Long}, which requires complex multi-step reasoning. Best results are \textbf{bolded}, and second-best are \underline{underlined}.}
\label{tab:libero_results}
\small
\setlength{\tabcolsep}{8pt}
\resizebox{.8\linewidth}{!}{
\begin{tabular}{lccccc}
\toprule
\textbf{Method} & \textbf{L-Spatial} & \textbf{L-Object} & \textbf{L-Goal} & \textbf{L-Long} & \textbf{Average} \\
\midrule
Diffusion Policy \citep{chi2025diffusion} & 78.5 & 87.5 & 73.5 & 64.8 & 76.1 \\
OpenVLA \citep{kim2024openvla} & 84.7 & 88.4 & 79.2 & 53.7 & 76.5 \\
SpatialVLA \citep{qu2025spatialvla} & 88.2 & 89.9 & 78.6 & 55.5 & 78.1 \\
CoT-VLA \citep{zhao2025cotvla} & 87.5 & 91.6 & 87.6 & 69.0 & 83.9 \\
$\pi_0$-Fast \citep{pertsch2025fast} & 96.4 & 96.8 & 88.6 & 60.2 & 85.5 \\
\midrule
GR00T-N1 \citep{nvidia2025gr00t} & 94.4 & 97.6 & 93.0 & 90.6 & 93.9 \\
$\pi_0$ \citep{black2024pi_0} & 98.0 & 96.8 & 94.4 & 88.4 & 94.4 \\
F1 \citep{lv2025f1} & 98.2 & 97.8 & 95.4 & 91.3 & 95.7 \\
InternVLA-M1 \citep{chen2025internvla-m1} & 98.0 & 99.0 & 93.8 & 92.6 & 95.9 \\
Discrete Diffusion VLA \citep{liang2025discrete} & 97.2 & 98.6 & 97.4 & 92.0 & 96.3 \\
$\pi_{0.5}$ \citep{intelligence2025pi05} & 98.8 & 98.2 & 98.0 & 92.4 & 96.9 \\
OpenVLA-OFT \citep{kim2025openvla-oft} & 97.6 & 98.4 & 97.9 & 94.5 & 97.1 \\
X-VLA \citep{zheng2025xvla} & 98.2 & 98.6 & 97.8 & \textbf{97.6} & 98.1 \\
EO1 \citep{qu2025eo1} & \textbf{99.7} & \textbf{99.8} & \underline{99.2} & 94.8 & \underline{98.2} \\
\midrule
\rowcolor{BlockA!20} \textbf{Being-H0.5}(generalist) & 97.0 & 98.2 & 99.0 & 96.2 & 97.6 \\
\rowcolor{BlockA!20} \textbf{Being-H0.5}(specialist) & \underline{99.2} & \underline{99.6} & \textbf{99.4} & \underline{97.4} & \textbf{98.9} \\
\bottomrule
\end{tabular}
}
\end{table}

We evaluate \textbf{Being-H0.5} on two widely-used simulation benchmarks, LIBERO~\cite{liu2023libero} and RoboCasa~\cite{nasiriany2024robocasa},
and compare against recent VLA systems and specialized robotic policies.
Unless otherwise specified, Being-H0.5 uses \textbf{RGB-only} observations at \textbf{224$\times$224} resolution with a \textbf{2B} backbone.
\textbf{Specialist vs.\ Generalist.}
Throughout this section, we similarly report two training regimes:
\textbf{(i) specialist}, which trains a benchmark-specific checkpoint using data from \emph{only one} benchmark (LIBERO-only or RoboCasa-only);
and \textbf{(ii) generalist}, which trains a \emph{single} checkpoint jointly on LIBERO + RoboCasa.
To ensure a fair comparison in terms of optimization, the generalist regime runs for 2$\times$ the training steps of a specialist model,
so that the model sees a comparable number of update steps \emph{per benchmark} while learning a unified policy across both testbeds.
At evaluation time, the same generalist checkpoint is directly evaluated on each benchmark without any additional adaptation.

\subsubsection{Results on LIBERO}
\label{sec:libero}

\paragraph{Experimental setup.}
We follow the standard LIBERO evaluation protocol~\cite{liu2023libero,kim2025fine}.
The policy takes multi-view RGB observations (wrist-mounted and third-person cameras), each resized to $224\times224$.
We use action chunking with chunk size 8 and train with packed sequences (7,680 tokens per GPU) for an effective batch size of 128.
Optimization runs for 45k steps on 4$\times$A800 GPUs for the specialist model.
For the generalist model, we train for approximately \textbf{2$\times$} steps under the joint LIBERO+RoboCasa mixture.
For evaluation, we run 500 trials per suite.

\paragraph{Results and analysis.}
Table~\ref{tab:libero_results} shows that Being-H0.5(\textbf{specialist}) reaches a \textbf{98.9\%} average success rate on LIBERO,
with strong performance across all suites including \texttt{LIBERO-Long} (97.4\%).
Being-H0.5(\textbf{generalist}), trained jointly on LIBERO+RoboCasa with 2$\times$ steps, remains highly competitive at \textbf{97.6\%} average.
The modest gap indicates that a single checkpoint can absorb substantial cross-benchmark diversity without sacrificing much LIBERO performance,
even though the generalist must allocate capacity to cover RoboCasa's long-horizon kitchen interactions in addition to LIBERO's task distribution.
Crucially, both results are achieved using only \textbf{224$\times$224 RGB} inputs and a \textbf{2B} backbone, matching or outperforming many higher-resolution, larger, or multi-modal systems.

\begin{table}[h]
\centering
\caption{\textbf{Success rates (\%) on the RoboCasa 24-task benchmark}. Being-H0.5 is evaluated across three task categories, averaged over 50 trials per task. Despite using only 224$\times$224 RGB inputs, Being-H0.5 outperforms state-of-the-art 3D-based methods and existing VLAs on the overall average. Best results are \textbf{bolded}, and second-best are \underline{underlined}.}
\label{tab:robocasa_results_refined}
\small
\setlength{\tabcolsep}{6pt}
\resizebox{.9\linewidth}{!}{
\begin{tabular}{llcccc}
\toprule
\textbf{Modality} & \textbf{Method} & \textbf{Pick \& Place} & \textbf{Doors/Drawers} & \textbf{Others} & \textbf{Total Avg.} \\
\midrule
\rowcolor{gray!5} \textit{3D (Multi-modal)} & & & & & \\
3D & 3DA \cite{3d_diffuser_actor} & 0.0 & 2.3 & 13.1 & 5.5 \\
3D & DP3 \cite{Ze2024DP3} & 1.5 & 41.7 & 32.0 & 22.8 \\
3D & GWM \cite{lu2025gwm} & 14.8 & 54.3 & 49.8 & 39.3 \\
\midrule
\rowcolor{gray!5} \textit{RGB-only (Standard)} & & & & & \\
RGB (256$\times$256) & BC \cite{nasiriany2024robocasa} & 4.3 & 47.0 & 42.2 & 28.9 \\
RGB (256$\times$256) & GR00T-N1 \cite{nvidia2025gr00t} & 18.6 & 50.2 & 39.1 & 36.0 \\
RGB (256$\times$256) & $\pi_{0.5}$ \cite{intelligence2025pi05} & 21.5 & 57.8 & 44.9 & 41.4 \\
RGB (256$\times$256) & $\pi_{0}$ \cite{black2024pi_0} & 14.0 & 53.1 & \textbf{58.5} & 42.4 \\
\midrule
\rowcolor{BlockA!20} RGB (224$\times$224) & \textbf{Being-H0.5}(generalist) & \textbf{40} & \textbf{73} & 52 & \underline{53.3} \\
\rowcolor{BlockA!20} RGB (224$\times$224) & \textbf{Being-H0.5}(specialist) & \underline{36} & \underline{71.7} & \underline{57.6} & \textbf{53.9} \\
\bottomrule
\end{tabular}
}
\end{table}

\subsubsection{Results on RoboCasa}
\label{sec:robocasa}

\paragraph{Experimental setup.}
We evaluate on RoboCasa~\cite{nasiriany2024robocasa}, which contains 24 long-horizon household tasks in diverse kitchen environments.
We adopt the challenging \textbf{Human-50} few-shot setting, using 50 human demonstrations per task.
Evaluation is performed with 50 trials per task across five held-out scenes, featuring unseen object instances and unseen kitchen styles.
We use \textbf{RGB-only} inputs at $224\times224$ and natural-language task commands, without depth or point clouds.
The specialist model is trained for the same budget as above (RoboCasa-only), while the generalist is trained jointly on LIBERO+RoboCasa for 2$\times$ steps.

\paragraph{Results and analysis.}
As shown in Table~\ref{tab:robocasa_results_refined}, Being-H0.5 achieves a new state of the art on RoboCasa:
\textbf{53.9\%} success for the specialist checkpoint and \textbf{53.3\%} for the generalist single-checkpoint model.
Notably, the generalist preserves essentially the same performance while simultaneously maintaining strong LIBERO performance,
suggesting that joint training does not introduce severe negative transfer between these benchmarks under our unified action interface.
The largest gain appears in Pick \& Place, where Being-H0.5 reaches 36--40\% success, substantially exceeding prior VLAs
(e.g., $\pi_{0.5}$ at 21.5\%) and even outperforming multi-modal baselines that leverage explicit 3D inputs.
This is particularly notable because RoboCasa pick-and-place demands reliable spatial reasoning and consistent grasp--transport--place coordination under appearance variation.
Our results indicate that large-scale human-centric pretraining can distill transferable spatial priors that remain effective even under RGB-only inputs and reduced resolution.

\paragraph{Takeaway.}
Across both benchmarks, the specialist regime provides an upper bound when benchmark-specific adaptation is allowed,
while the generalist regime shows that a \textbf{single} Being-H0.5 checkpoint can jointly cover LIBERO and RoboCasa with only a small degradation on LIBERO and near-identical performance on RoboCasa.
This points to a practical path toward scalable deployment where maintaining many benchmark-specific checkpoints is undesirable.

\begin{table*}[t]
\centering
\caption{\textbf{Impact of human-centric learning in single-task 5-shot adaptation.} 
We investigate this by selectively freezing key architectural components.
\textbf{Native VLM}: Initialized from the original large multimodal model. 
\textbf{Human-Centric}: Initialized with our human-centric pretrained weights.
The action expert remains trainable in all settings.
$\Delta$ denotes the absolute performance gain.
}
\label{tab:abl_pretrain_single}
\small
\setlength{\tabcolsep}{9pt}
\begin{tabular}{ll ccccc}
\toprule
\multicolumn{2}{c}{\textbf{Model Configuration}} & \multicolumn{5}{c}{\textbf{Single-Task LIBERO 5-Shot Success Rate (\%)}} \\
\cmidrule(lr){1-2} \cmidrule(lr){3-7}
\textbf{Frozen Components} & \textbf{Initialization} & \textbf{L-Spatial} & \textbf{L-Object} & \textbf{L-Goal} & \textbf{L-Long} & \textbf{Avg.} \\
\midrule

\multirow{3}{*}{\textbf{Und + Proj + ViT}} 
    & Native VLM & 67.8 & 76.8 & 57.2 & 29.8 & 57.9 \\
    & Human-Centric & 78.6 & 78.9 & 72.6 & 46.2 & 69.0 \\
    \rowcolor{BlockA!30}
    & \textbf{Gain ($\Delta$)} & \textbf{+10.8} & \textbf{+2.1} & \textbf{+15.4} & \textbf{+16.4} & \textbf{+11.1} \\
\addlinespace[0.6em]

\multirow{3}{*}{\textbf{Und + ViT}}       
    & Native VLM & 70.6 & 50.0 & 67.0 & 17.4 & 51.3 \\
    & Human-Centric & 85.0 & 84.2 & 80.0 & 59.0 & 77.1 \\
    \rowcolor{BlockA!30}
    & \textbf{Gain ($\Delta$)} & \textbf{+14.4} & \textbf{+34.4} & \textbf{+13.0} & \textbf{+41.6} & \textbf{+25.8} \\
\addlinespace[0.6em]
\midrule 

\multirow{3}{*}{\textbf{Und Only}}             
    & Native VLM & 78.4 & 84.8 & 62.0 & 49.2 & 68.6 \\
    & Human-Centric & 83.2 & 79.6 & 74.0 & 63.6 & 75.1 \\
    \rowcolor{BlockA!30}
    & \textbf{Gain ($\Delta$)} & \textbf{+4.8} & \textcolor{red}{-5.2} & \textbf{+12.0} & \textbf{+14.4} & \textbf{+6.5} \\
\addlinespace[0.6em]

\multirow{3}{*}{\textbf{Proj + ViT}}      
    & Native VLM & 87.2 & 86.6 & 80.8 & 58.0 & 78.2 \\
    & Human-Centric & 86.2 & 86.2 & 84.6 & 66.8 & 81.0 \\
    \rowcolor{BlockA!30}
    & \textbf{Gain ($\Delta$)} & \textcolor{red}{-1.0} & \textcolor{red}{-0.4} & \textbf{+3.8} & \textbf{+8.8} & \textbf{+2.8} \\
\addlinespace[0.6em]

\multirow{3}{*}{\textbf{None (Full FT)}}  
    & Native VLM & 84.2 & 83.6 & 75.0 & 65.8 & 77.2 \\
    & Human-Centric & 87.2 & 91.4 & 81.6 & 66.8 & 81.8 \\
    \rowcolor{BlockA!30}
    & \textbf{Gain ($\Delta$)} & \textbf{+3.0} & \textbf{+7.8} & \textbf{+6.6} & \textbf{+1.0} & \textbf{+4.6} \\
\bottomrule
\end{tabular}
\end{table*}

\begin{table*}[t]
\centering
\caption{\textbf{Impact ablation of human-centric learning in multi-task 5-shot adaptation.} 
In contrast to the single-task setup, the model is jointly trained on data from all four task suites and subsequently evaluated on each individually. 
\textbf{Native VLM}: Initialized from the original large multimodal model. 
\textbf{Human-Centric}: Initialized with our human-centric pretrained weights.
This multi-task protocol serves to rigorously assess the generalization capabilities and robustness of VLAs under diverse cross-task scenarios.}
\label{tab:abl_pretrain_multi}
\small
\setlength{\tabcolsep}{9pt} 
\begin{tabular}{ll ccccc}
\toprule
\multicolumn{2}{c}{\textbf{Model Configuration}} & \multicolumn{5}{c}{\textbf{Multi-task-suite LIBERO 5-Shot Success Rate (\%)}} \\
\cmidrule(lr){1-2} \cmidrule(lr){3-7}
\textbf{Frozen Components} & \textbf{Initialization} & \textbf{L-Spatial} & \textbf{L-Object} & \textbf{L-Goal} & \textbf{L-Long} & \textbf{Avg.} \\
\midrule

\multirow{3}{*}{\textbf{Und + Proj + ViT}} 
    & Native VLM & 61.6 & 77.0 & 68.0 & 36.0 & 60.7 \\
    & Human-Centric & 80.8 & 81.8 & 74.4 & 52.4 & 72.4 \\
    \rowcolor{BlockA!30}
    & \textbf{Gain ($\Delta$)} & \textbf{+19.2} & \textbf{+4.8} & \textbf{+6.4} & \textbf{+16.4} & \textbf{+11.7} \\
\addlinespace[0.6em]

\multirow{3}{*}{\textbf{Und + ViT}}       
    & Native VLM & 78.4 & 84.8 & 71.4 & 38.6 & 68.3 \\
    & Human-Centric & 83.6 & 82.6 & 80.8 & 59.2 & 76.6 \\
    \rowcolor{BlockA!30}
    & \textbf{Gain ($\Delta$)} & \textbf{+5.2} & \textcolor{red}{-2.2} & \textbf{+9.4} & \textbf{+20.6} & \textbf{+8.3} \\
\addlinespace[0.6em]
\midrule

\multirow{3}{*}{\textbf{Und Only}}             
    & Native VLM & 75.8 & 84.2 & 71.2 & 59.8 & 72.8 \\
    & Human-Centric & 85.2 & 90.6 & 84.2 & 68.4 & 82.1 \\
    \rowcolor{BlockA!30}
    & \textbf{Gain ($\Delta$)} & \textbf{+9.4} & \textbf{+6.4} & \textbf{+13.0} & \textbf{+8.6} & \textbf{+9.3} \\
\addlinespace[0.6em]

\multirow{3}{*}{\textbf{Proj + ViT}}      
    & Native VLM & 87.6 & 91.0 & 82.2 & 62.6 & 80.9 \\
    & Human-Centric & 89.6 & 88.8 & 89.6 & 65.8 & 83.5 \\
    \rowcolor{BlockA!30}
    & \textbf{Gain ($\Delta$)} & \textbf{+2.0} & \textcolor{red}{-2.2} & \textbf{+7.4} & \textbf{+3.2} & \textbf{+2.6} \\
\addlinespace[0.6em]

\multirow{3}{*}{\textbf{None (Full FT)}}  
    & Native VLM & 87.8 & 94.6 & 82.6 & 71.4 & 84.1 \\
    & Human-Centric & 88.8 & 90.2 & 85.4 & 76.0 & 85.1 \\
    \rowcolor{BlockA!30}
    & \textbf{Gain ($\Delta$)} & \textbf{+1.0} & \textcolor{red}{-4.4} & \textbf{+2.8} & \textbf{+4.6} & \textbf{+1.0} \\
\bottomrule
\end{tabular}
\end{table*}

\begin{figure}[ht] 
\centering
\includegraphics[width=.95\linewidth]{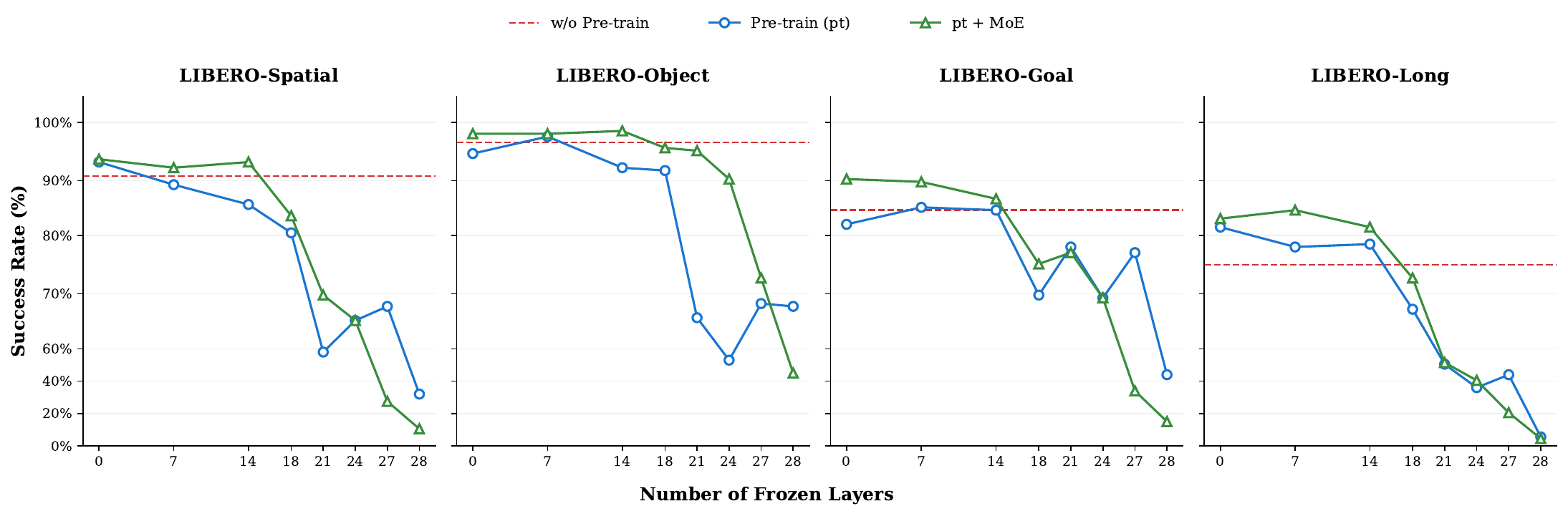}
 \caption{\textbf{Ablation study on the number of frozen layers in the action expert.} Results compare the performance of the model after human-centric pretraining (``pt'') and without it (``w/o pt''). ``MoF'' denotes our proposed Mixture of Flow architecture. All experiments freeze all parameters of the MLLM including the visual encoder and projector. Note that each task was evaluated with 20 rollouts due to time constraints.}
\label{fig:abl_freeze_expert}
\end{figure}

\subsection{Ablation Study}

\subsubsection{How Do Human-Centric Learning Benefit Downstream Adaptation?}

\begin{figure}[ht] 
\centering
\includegraphics[width=.95\linewidth]{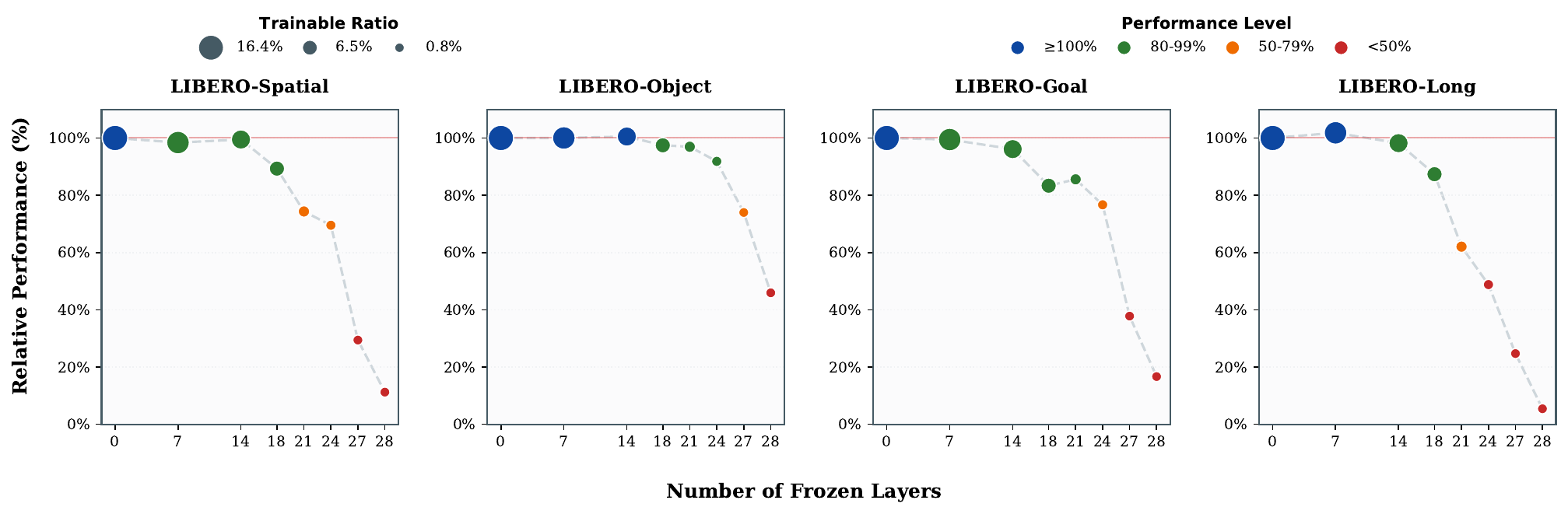}
 \caption{\textbf{Ablation Study on the Number of Frozen Layers in the Action Expert.} Results compare the performance of the model after human-centric pretraining (``pt'') and without it (``w/o pt''). ``MoF'' denotes our proposed Mixture of Flow architecture. All experiments freeze all parameters of the MLLM, including the visual encoder and projector. Note that each task was evaluated with 20 rollouts due to time constraints.}
\label{fig:abl_freeze_expert_relative}
\end{figure}

Our human-centric learning provides foundational priors that facilitate efficient adaptation to downstream robotic tasks.
In this section, we analyze the specific impact of these pretrained weights on the adaptation process.
To mitigate the risk of overfitting, which is a prevalent challenge in robotic benchmarks, and to rigorously evaluate few-shot generalization, we adopt the LIBERO 5-shot benchmark with a restricted 10K-step training window.
By utilizing a minimal set of training samples, we can effectively isolate performance gains directly attributable to pretrained knowledge versus task-specific rote learning.
Our validation follows a dual-protocol approach:
\textbf{1) Singe-task-suite:} Training separate models for each LIBERO suite to observe pretrained effectiveness in singular task scenarios.
\textbf{2) Multi-task-suite:} Training a unified model across all suites to assess the capability of the pretrained VLA to adapt to large-scale, diverse task distributions.
The results are detailed in Table \ref{tab:abl_pretrain_single} and Table \ref{tab:abl_pretrain_multi}.
In these tables, we compare the performance of models initialized with our human-centric weights against a baseline initialized directly from the native VLM. 
Note that the Action Expert remains trainable in all experimental configurations.
Additionally, Figures \ref{fig:abl_freeze_expert} and \ref{fig:abl_freeze_expert_relative} further explore the sensitivity of the Action Expert to parameter freezing.

\paragraph{Impact of Backbone Pretraining.}
The results reveal critical insights into how weight initializations influence VLA performance across different paradigms.
First, the impact of pretraining is most pronounced under maximum parameter constraints.
As shown in Table \ref{tab:abl_pretrain_single}, when the core reasoning components (Und + ViT) are frozen while allowing the projector and action expert to train, the pretrained model achieves a remarkable \textbf{+25.8\%} average gain over the baseline.

A consistent trend across both single- and multi-task setups is that LIBERO-Long tasks benefit disproportionately from pretraining.
In the single-task setup, freezing the Und + ViT components yields a staggering +41.6\% improvement in the Long suite. 
This indicates that while simple spatial or object-oriented tasks can be learned from scratch with 50 demonstrations, the temporal consistency and intent-grounding required for long-horizon manipulation are almost entirely derived from the foundational ``world knowledge" embedded in the pretrained VLM.
Interestingly, the marginal utility of pretraining ($\Delta$) decreases as the model gains more trainable parameters (moving toward Full FT).
In some cases, such as L-Object, we observe slight negative transfer (e.g., -4.4\% in Multi-task Full FT).
This suggests that for simpler, object-centric tasks, aggressive joint fine-tuning may lead the model to ``overwrite" specialized pretrained features in favor of task-specific heuristics.
Our ablation confirms that the Language (Und) and Visual (ViT) backbones act as the primary carriers of generalizable robotic proficiency, providing a stable ``intent manifold'' that prevents overfitting when downstream data is sparse.

\paragraph{Sensitivity of Action Expert Plasticity.}
While the backbone benefits from stability, the Action Expert requires plasticity.
We investigate the impact of freezing layers within the Action Expert, ranging from the input to the output heads, as visualized in Figure \ref{fig:abl_freeze_expert}.
The results reveal that freezing the initial 0 to 7 layers results in negligible performance degradation, as the model sustains robust success rates exceeding 80\% across the majority of task suites.
However, as the number of frozen layers exceeds 14, performance degrades sharply, dropping below 20\% when the entire expert is frozen.
Figure \ref{fig:abl_freeze_expert_relative} highlights this trade-off between trainable parameters and performance relative to the peak.
This indicates that while the high-level semantic planning is transferable from human-centric pretraining, the low-level motor control—mapping features to precise joint velocities—must be actively learned from downstream demonstrations.
Therefore, the optimal configuration for low-data adaptation is to freeze the pretrained semantic backbone (Und + ViT) while keeping the Action Expert fully trainable.

\subsubsection{How Does Masked Motion Token Prediction Benefit Capturing Behavior Priors?}
We further verify the benefit of introducing the discrete \textit{Masked Motion Token Prediction} objective in our pre-training.
Specifically, we conduct an ablation that removes $\mathcal{L}_{\textsc{mask}}$ while keeping all other settings unchanged, and evaluate motion generation on a held-out subset of our human demonstration data.
We split the evaluation set by data source into two domains: (i) \textbf{lab-curated} demonstrations collected under controlled setups, and (ii) \textbf{in-the-wild} videos featuring diverse scenes and heterogeneous motion patterns.

\begin{wraptable}{r}{0.4\linewidth}
\centering
\caption{\textbf{Ablation on masked motion token prediction.} MWDS $\uparrow$ on a held-out human demonstration test set with \textbf{Lab} (lab-curated) and \textbf{Wild} (in-the-wild) splits.}
\label{tab:ablation_mask_mwds}
\vspace{-2mm}
\begin{tabular}{lcc}
\toprule
Method & Lab $\uparrow$ & Wild $\uparrow$ \\
\midrule
Hybrid (Ours) & 0.33 & 0.20  \\
w/o $\mathcal{L}_{\textsc{mask}}$ & \textbf{0.35} & \textbf{0.28} \\
\bottomrule
\end{tabular}
\vspace{-4mm}
\end{wraptable}

Notably, since our unified action space adopts a \emph{delta} representation, typical position error metrics such as Mean Per Joint Position Error (MPJPE) become less informative: the initial pose usually dominates such errors, while the initial state in our setting is provided as ground truth.
Therefore, we instead measure whether the model captures the \emph{intent} in language and grounds it in the image geometry using \textbf{Mean Wrist Displacement Similarity (MWDS)}.
Concretely, MWDS averages the cosine similarity between the predicted and ground-truth wrist displacement vectors from the initial to the final state,
$\cos\!\left(\Delta\hat{\mathbf{p}}_{\text{wrist}},\,\Delta\mathbf{p}_{\text{wrist}}\right)$.

As shown in Table~\ref{tab:ablation_mask_mwds}, removing masked prediction leads to a clear drop in MWDS on both the lab-curated and in-the-wild splits.
This indicates that the discrete masked objective injects a stable, abstraction-level behavior prior that improves intention-aligned generation and enhances robustness across heterogeneous motion patterns. Notably, the improvement is more pronounced on the \textit{Wild} split, where demonstrations contain substantially higher behaviour variability and observation noise. This suggests that the abstraction-level masked objective particularly matters for noisy, large-scale human-centric pre-training.

\begin{figure}[ht]
\centering
\includegraphics[width=.8\linewidth]{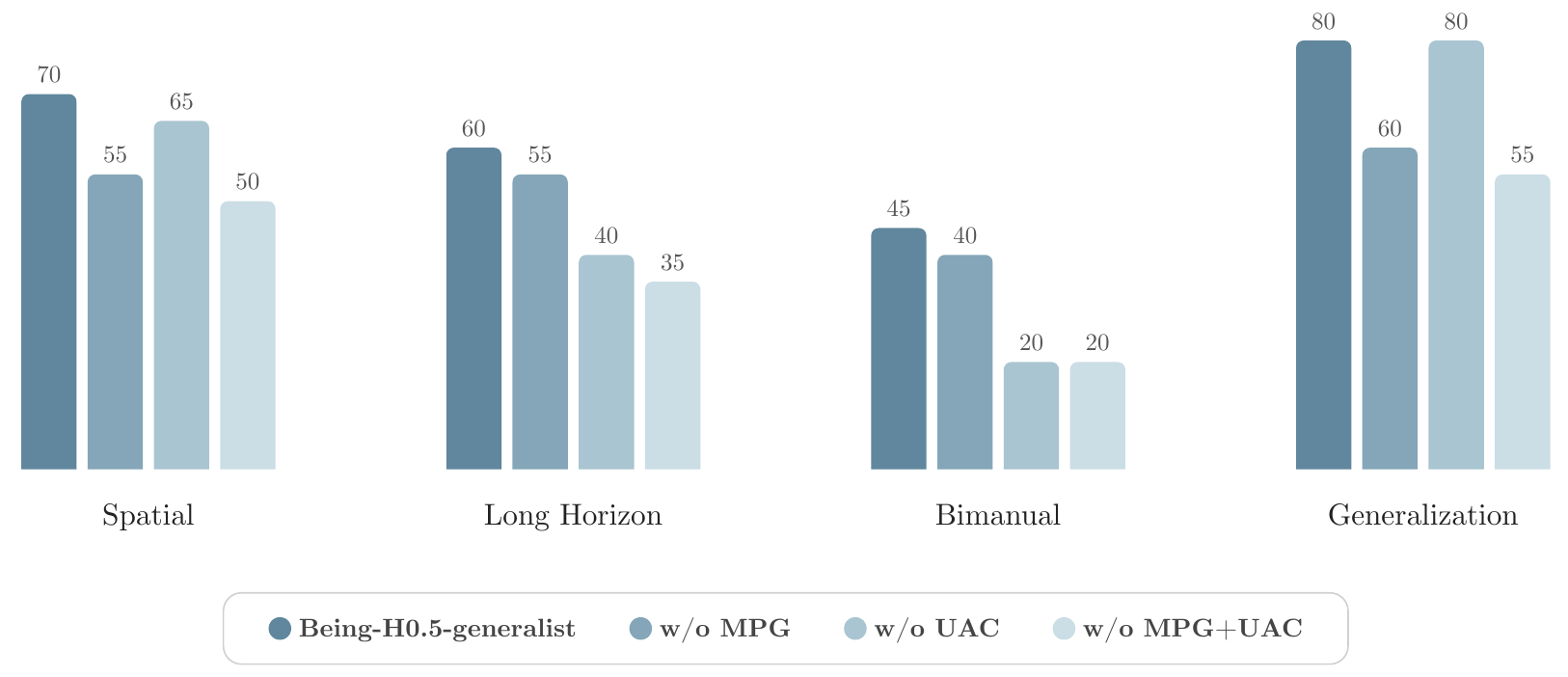}
\caption{\textbf{Ablation of MPG+UAC on real robots.} Removing MPG+UAC hurts long-horizon and bimanual categories the most, where execution delay and unreliable context amplify compounding errors.}
\label{fig:exp_real_bars_mpg_uac}
\end{figure}

\subsubsection{Real-Time Efficiency: MPG and UAC}
\label{sec:real_robot_mpg_uac_ablation}

We further ablate the deployment-time components by disabling both \textbf{MPG} (manifold-preserving gating) and \textbf{UAC} (universal async chunking),
while keeping the backbone and training data fixed.
Figure~\ref{fig:exp_real_bars_mpg_uac} shows consistent degradations.

\paragraph{Long-horizon drops sharply without UAC-style async consistency.}
Long-horizon tasks involve multi-stage behaviors with execution delays and asynchronous sensing/actuation.
Without UAC, the policy becomes more sensitive to temporal mismatch between perceived state and the committed portion of the action chunk,
leading to error accumulation (e.g., drifting grasps, premature contact, failed handle engagement) and a noticeable drop in success.

\paragraph{Bimanual coordination is brittle without stability mechanisms.}
Bimanual tasks require tight temporal coupling between two effectors under partial observability and contact.
Disabling MPG+UAC increases oscillation and indecisive corrections when the observation context is noisy or changes rapidly,
especially for platforms with movable cameras where viewpoint shifts can invalidate short-term visual cues.

\paragraph{Spatial/generalization are affected but less catastrophically.}
Spatial tasks still rely on precise placement, yet the execution horizon is shorter, and there are fewer compounding subgoals.
Generalization tasks benefit from robustness mechanisms when the scene shifts, but typically allow more tolerance and recovery, hence the smaller (yet consistent) degradation.

\section{Conclusion}

In this work, we present Being-H0.5, a foundational Vision-Language-Action (VLA) model that establishes a scalable, human-centric paradigm for cross-embodiment robot learning. By unifying human and robotic motion within a physically grounded state-action space and leveraging the large-scale UniHand-2.0 dataset, we demonstrate that human interaction traces can serve as dense physical priors to bridge the gap between diverse robotic morphologies. Our Mixture-of-Transformers (MoT) architecture effectively disentangles high-level multimodal reasoning from low-level motor execution, while Manifold-Preserving Gating (MPG) and Universal Async Chunking (UAC) ensure robust deployment by mitigating contextual distribution shifts and hardware-specific inference latencies. Extensive evaluation across a heterogeneous fleet of five robotic platforms proves that a single Being-H0.5 checkpoint can internalize complex interaction dynamics, achieving high success rates in bimanual coordination and long-horizon tasks. Ultimately, Being-H0.5 signifies a transition toward versatile, general-purpose robotic intelligence capable of seamless operation across various environments and embodiments.

\clearpage

\bibliographystyle{unsrt}
\bibliography{ref}

\clearpage

\beginappendix
\section{Contributions}

$\dagger$ denotes leading the part


\begin{multicols}{2} 

\noindent\textbf{Core Contributors (alphabetical order)}

\begin{itemize}
    \item \textsc{Hao Luo}
    \item \textsc{Ye Wang}
    \item \textsc{Wanpeng Zhang}
    \item \textsc{Sipeng Zheng}
\end{itemize}

\noindent\textbf{Model Training}
\begin{itemize}
    \item \textsc{Hao Luo}
    \item \textsc{Ye Wang}
    \item \textsc{Wanpeng Zhang}
    \item \textsc{Sipeng Zheng}
\end{itemize}

\noindent\textbf{Infrastructure}
\begin{itemize}
    \item \textsc{Wanpeng Zhang}$^{\dagger}$
    \item \textsc{Ye Wang}
    \item \textsc{Chaoyi Xu}
    \item \textsc{Chi Zhang}
    \item \textsc{Haoqi Yuan}
\end{itemize}

\noindent\textbf{Human Data}
\begin{itemize}
    \item \textsc{Hao Luo}$^{\dagger}$
    \item \textsc{Ziheng Xi}
    \item \textsc{Yiqing Wang}
    \item \textsc{Yicheng Feng}
\end{itemize}

\noindent\textbf{Robot Data}
\begin{itemize}
    \item \textsc{Ye Wang}$^{\dagger}$
    \item \textsc{Sipeng Zheng}$^{\dagger}$
    \item \textsc{Haiweng Xu}
\end{itemize}

\columnbreak

\noindent\textbf{Experiments}
\begin{itemize}
    \item \textsc{Wanpeng Zhang}
    \item \textsc{Ye Wang}
    \item \textsc{Hao Luo}
    \item \textsc{Sipeng Zheng}
\end{itemize}

\noindent\textbf{Project Lead}
\begin{itemize}
    \item \textsc{Sipeng Zheng}
    \item \textsc{Zongqing Lu}
\end{itemize}

\end{multicols} 

\clearpage

\section{Full Rollouts}
\label{sec:full_rollouts}

\begin{figure}[ht]
\centering
\includegraphics[width=.96\linewidth]{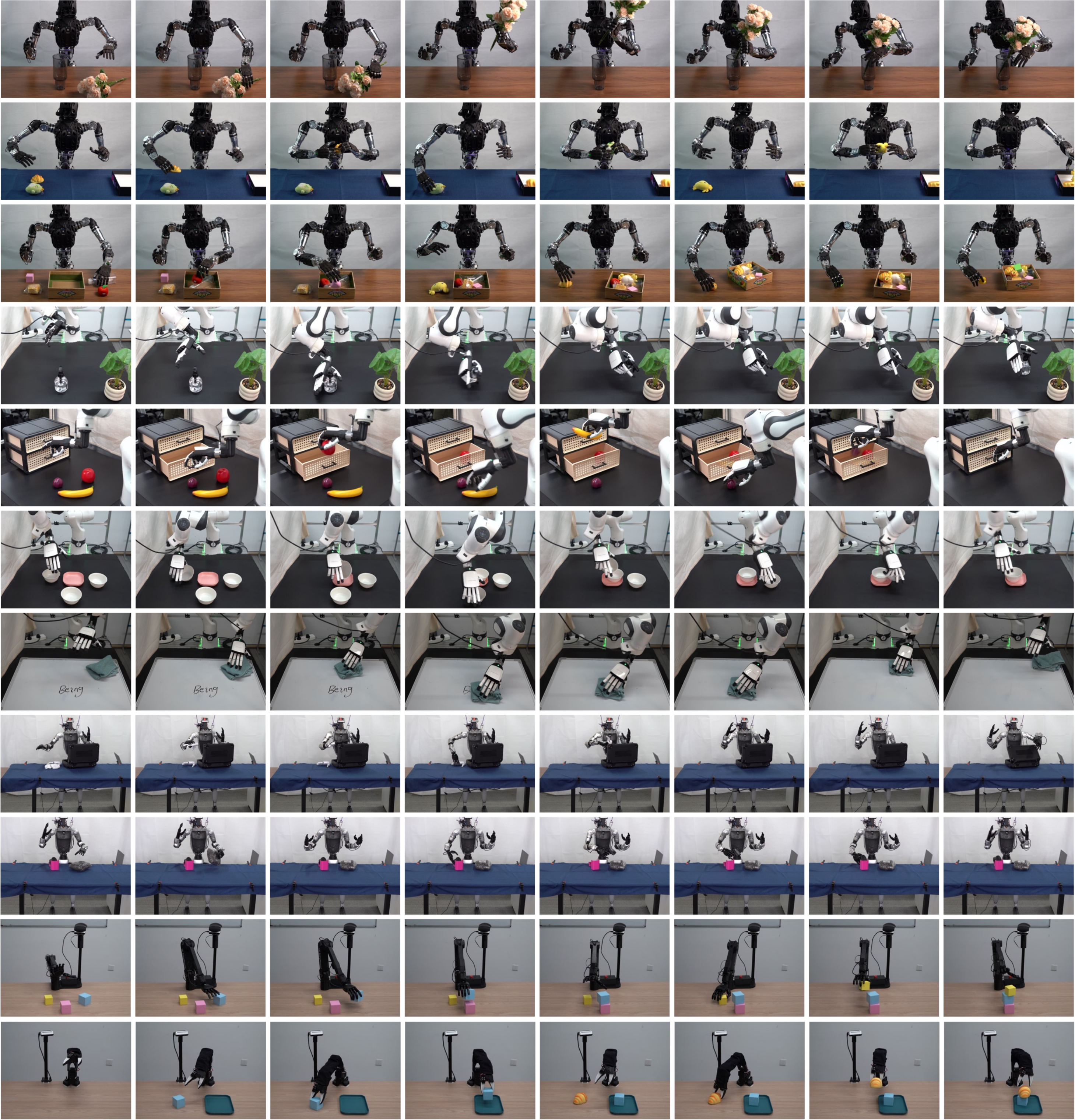}
\caption{\textbf{Example rollouts of all real-robot task suites.}}
\label{fig:all_rollouts}
\end{figure}

\end{document}